\pdfoutput=1

\documentclass[opre,nonblindrev]{informs3_homepage}

\usepackage{natbib}
\bibpunct[, ]{(}{)}{,}{a}{}{,}%
\def\bibfont{\small}%

\usepackage{microtype}

\usepackage{suffix}
\usepackage{grffile}

\usepackage{subcaption}

\usepackage{algorithm}
\usepackage{algorithmic}

\usepackage[informs]{autoconfig}

\DeclareMathAlphabet\mathbfcal{OMS}{cmsy}{b}{n}

\def\Indicator{\mathbb{I}}

\def\LinTS{LinTS}
\def\TSAI{TS-AI}

\def\Policy{\pi}
\def\PolicyTS{\Policy^{\operatorname{\LinTS}}}

\def\Arm{A}

\def\ArmSet{\mathcal{\Arm}}

\def\ArmBound{\mathbf{a}}

\def\ChosenArm{\widetilde{\Arm}}
\def\OptimalArm{\Arm^\star}

\def\Regret{\operatorname{Regret}}
\def\History{\mathcal{F}}
\def\HistoryPlus{\History}

\def\Param{\Theta^\star}

\def\GainRate{\mathsf{G}}
\WithSuffix\def\GainRate*[#1]{\GainRate_{#1}}

\def\GapProb{\mathsf{q}}
\WithSuffix\def\GapProb*[#1]{\GapProb_{#1}}
\def\Deviation{\mathsf{D}}
\WithSuffix\def\Deviation*[#1]{\Deviation_{#1}}

\def\OptProb{\mathsf{p}}

\def\MeanReward{\mathsf{M}}

\def\Worth{\widetilde{\MeanReward}}
\def\ITypical{\mathbb{T}}
\def\ITypicalM{\ITypical^\star}

\def\eps{\varepsilon}

\def\Inflation{\iota}
\def\TsSample{\widetilde{\Theta}}
\def\ParamBound{\boldsymbol{\theta}}

\def\DivParam{\nu}

\def\IOptim{\mathbb{O}}
\def\RewardValue{Y}
\def\Thinness{\psi}
\def\ExploreRatio{\omega}
\def\ThinnessBound{\Psi}
\def\ConfSet{\mathcal{C}}
\def\Error{E}
\def\Compensator{C}

\def\BiasCoeff{\beta}
\def\Prior{\mathcal{P}_{\Param}}
\def\BadEvent{\mathcal{B}}
\def\One{\mathbf{1}}

\def\Eye{\mathbb{I}}
\def\Dim{d}

\def\SymCovMatrix{\mathbf{\Sigma}}
\def\Estimator{\widehat{\Theta}}
\def\Radius{\rho}
\def\RadiusParam{\Radius^\star}
\def\RadiusWorth{\widetilde{\Radius}}

\def\Uniform{\operatorname{Unif}}

\def\II{\mathbb{I}}
\def\IWellPosed{\mathbb{W}}

\def\randombias{randomization}
\newcommand\IS{\mathbb{S}}

\date{\today}
\usepackage{times}

\ECRepeatTheorems

\begin{document}
	
	
	
	\RUNTITLE{Frequentist Regret of Thompson Sampling}
	
	\TITLE{On Frequentist Regret of Linear Thompson Sampling}
	
	
	\ARTICLEAUTHORS{%
		\AUTHOR{Nima Hamidi}
		\AFF{Department of Statistics, Stanford University, \EMAIL{hamidi@stanford.edu}}
		\AUTHOR{Mohsen Bayati}
		\AFF{
			Graduate School of Business, Stanford University, \EMAIL{bayati@stanford.edu}}
	} 
	
	\ABSTRACT{This paper studies the stochastic linear bandit problem, where a decision-maker chooses actions from possibly time-dependent sets of vectors in $\mathbb{R}^d$ and receives noisy rewards. The objective is to minimize regret, the difference between the cumulative expected reward of the decision-maker and that of an oracle with access to the expected reward of each action, over a sequence of $T$ decisions. Linear Thompson Sampling (\LinTS{}) is a popular Bayesian heuristic, supported by theoretical analysis that shows its Bayesian regret is bounded by $\widetilde{\mathcal{O}}(d\sqrt{T})$, matching minimax lower bounds.
		However, previous studies demonstrate that the frequentist regret bound for \LinTS{} is $\widetilde{\mathcal{O}}(d\sqrt{dT})$, which requires posterior variance inflation and is by a factor of $\sqrt{d}$ worse than the best optimism-based algorithms. We prove that this inflation is fundamental and that the frequentist bound of $\widetilde{\mathcal{O}}(d\sqrt{dT})$ is the best possible, by demonstrating a \randombias{} bias phenomenon in \LinTS{} that can cause linear regret without inflation.
		We propose a data-driven version of \LinTS{} that adjusts posterior inflation using observed data, which can achieve minimax optimal frequentist regret, under additional conditions. Our analysis provides new insights into \LinTS{} and settles an open problem in the field.}

	
	
	\maketitle

\section{Introduction}
\label{sec:intro}

In recent years, an increasing number of organizations across diverse domains, including but not limited to e-commerce and digital advertising, are embracing the use of online experiments to optimize their decision-making process. However, conducting such experiments involves an opportunity cost, also known as \emph{regret}, caused by exposing some customers to potentially inferior experiences. To reduce this opportunity cost, a growing number of enterprises are turning to multi-armed bandit (MAB) experiments \citep{scott2010modern,scott2015multi,johari2017peeking}. The MAB approach works by adaptively utilizing the experiment's partially available results and favoring decisions with higher predicted value, or \emph{reward},
thus reducing their regret. The practical motivations for MAB problems, 
combined with their mathematical richness, have made them the subject of intense study in computer science, economics, operations research, and statistics \citep{bubeck2012regret,russo2018tutorial,lattimore2019bandit,slivkins2019introduction}.

This paper aims to answer an open question about one of the key algorithms used in MAB problems, which dates back to \cite{thompson1933likelihood}, in a general setting known as the stochastic linear bandit problem with changing action sets. This class includes the standard $k$-armed bandit problem, as well as the $k$-armed contextual bandit problem as special cases. In this setting, a decision-maker sequentially selects actions from given action sets and observes the corresponding rewards. The actions, which are vectors in $\IR^d$, can also be thought of as features or context that influence the rewards. The rewards are stochastic and their expectations depend on the actions through a fixed linear function, with an unknown parameter $\Param\in\IR^d$. As more decisions are made and their rewards are observed, the reward function can be estimated. The main objective of the decision-maker is to maximize the cumulative expected reward over a sequence of decision epochs. Alternatively, one can measure the expected regret or simply regret, which is the difference between the best achievable cumulative expected reward, obtained by an \emph{oracle} with access to the true expectation of the reward function, and the cumulative expected reward obtained by the decision-maker.

Regret can be measured in either a Bayesian or frequentist fashion. Bayesian regret is used when the unknown parameter $\Param$ is random, and the expectations are taken with respect to three sources: (1) the randomness in the reward functions, (2) the unknown parameter $\Param$, and (3) possible randomness introduced by the decision-maker. On the other hand, frequentist regret (also referred to as worst-case regret) is used when the parameter $\Param$ is deterministic, and the expectation is only with respect to the (1) and (3).



The main challenge faced by decision-makers is overcoming the \emph{curse of underestimation}, where the reward of the optimal action is underestimated, leading to its permanent discarding. To address this challenge, two approaches have gained considerable attention. The first approach, proposed by \cite{dani2008stochastic,rusmevichientong2010linearly} and improved by \cite{abbasi2011improved}, utilizes \emph{optimism in the face of uncertainty} (based on the Upper Confidence Bound technique due to \cite{lai1985asymptotically}), and obtains policies with frequentist regret bounds of $\OrderLog[\big]{d\sqrt{T}}$. \footnote{The notation $\OrderLog[\big]{.}$ is defined in \cref{sec:setting}.} As shown by \citet{dani2008stochastic}, this approach is minimax optimal up to logarithmic factors. The second approach, introduced by \cite{thompson1933likelihood}, arises from a Bayesian heuristic which suggests sampling from the posterior distribution of the reward function, given past observations, and choosing the best action as if this sample were the true reward function. This approach is known as Thompson sampling (TS) or posterior sampling, and although it is Bayesian in nature, it can be applied in the frequentist setting as well. TS is popular in practice due to its simplicity and good empirical performance, as reported by \cite{scott2010modern,scott2015multi,russo2018tutorial}.

TS has been extensively studied from a theoretical perspective. For the stochastic linear bandit problem, where the TS heuristic is referred to as \LinTS{}, \cite{russo2014learning} established a connection between \LinTS{} and optimistic policies and obtained a Bayesian regret bound of $\OrderLog[\big]{d\sqrt{T}}$, which is minimax optimal. However, in the frequentist setting, \cite{agrawal2013thompson} and \cite{abeille2017linear} have derived regret bounds of $\OrderLog[\big]{d\sqrt{dT}}$ for a variant of \LinTS{} (referred to as frequentist \LinTS{}) that samples from a posterior distribution with an \emph{inflated} variance of a factor $\OrderLog{d}$. 
When $d$ is not a constant, which is often the case in modern applications with a large number of customer-specific data that allows \emph{personalizing} the decisions, this bound is far from optimal, being worse by a factor of $\sqrt{d}$. While it is known in the literature that frequentist \LinTS{} has poor empirical performance due to its conservative over-exploration, it has remained an open question as to whether the inflation is necessary and whether the extra factor for the frequentist regret can be eliminated, see, for instance, \cite[\S8.1.2]{russo2018tutorial}. Our \emph{main contribution} in this paper is to answer this question negatively. In particular, we construct two families of examples to show that \LinTS{} without inflation suffers from a  \emph{\randombias{} bias} phenomenon and can incur linear regret when $d$ grows at least logarithmically in $T$ ($d=\LowerOrder{\log T}$) and at least one of the noise distribution or the prior distribution does not match the one that \LinTS{} assumes.

While the primary focus of this paper is theoretical, the examples we provide offer valuable insights into the inner workings of TS that could prove beneficial for practitioners. In practice, the prior and noise distributions are often unknown or difficult to sample from, necessitating the estimation or approximation of the posterior distribution. However, we demonstrate that even minor discrepancies between the true distributions and their estimates can significantly degrade the performance of TS, a problem that persists even when with noiseless reward function.

This shortcoming of TS can make certain applications vulnerable to adversarial attacks. Specifically, a commonly used assumption in posterior computation, that the set of actions is independent of the true reward function given past observations, can be violated if an adversary with partial knowledge of the true reward function can manipulate the action sets. For instance, consider an established marketplace platform $\mathcal{P}_1$ with extensive data that is competing with a new platform $\mathcal{P}_2$ that has limited access to data. $\mathcal{P}_2$ may use MAB experiments to expedite its learning and decision-making, but $\mathcal{P}_1$, with superior knowledge of the true reward function, could participate maliciously in $\mathcal{P}_2$'s marketplace via intermediary agents and diminish $\mathcal{P}_2$'s experimental performance. As we will show in \cref{sec:brittleness}, this malicious participation need not entail active monitoring of $\mathcal{P}_2$ and can be passively planned in advance, making it a plausible concern.

It is worth highlighting that optimism-based algorithms, such as the OFUL algorithm proposed by \cite{abbasi2011improved}, do not suffer from the aforementioned \randombias{} bias issue. Building on this insight, we aim to delve into the root cause of the problem with \LinTS{} and present a solution in the form of the \TSAI{} algorithm. Like \LinTS{}, \TSAI{} samples from the posterior distribution; however, it adapts the posterior variance based on the observed data, hence the acronym TS-AI, and only inflates it when additional exploration is necessary. This makes \TSAI{} less prone to over-exploration, while also eliminating the \randombias{} bias that plagues \LinTS{}. We establish that under additional assumptions, \TSAI{} achieves the frequentist minimax optimal regret. We also present numerical simulations that illustrate the limitations of \LinTS{}, and how \TSAI{} addresses them. These simulations confirm the advantages of \TSAI{} and how it retains the benefits of \LinTS{} while overcoming its limitations.

\subsection{Other related literature} 

In the special case of standard MAB problem, TS has been extensively studied, and several research works have established regret bounds for TS that match the minimax lower bounds up to logarithmic factors. \cite{agrawal2012analysis} and \cite{agrawal2013further} provide frequentist regret bounds for TS, while \cite{bubeck2013prior} demonstrate that TS attains optimal Bayesian regret up to constants. Recently, \cite{jin2020mots} proposed a modified version of TS that achieves minimax-optimal frequentist regret up to constant terms. 
In this domain, \cite{phan2019thompson} and \cite{nie2018adaptively} revealed interesting observations about TS that are related to our work. The former noted that sampling from an approximation of the posterior distribution with constant $\alpha$-divergence approximation error could result in linear regret. The latter demonstrated that the estimates for the mean rewards have a downward bias when a wide range of bandit algorithms collect the samples. Additionally, the recent work of \cite{bastani2019meta} provides a positive result for TS with misspecified prior in a dynamic pricing setting with a large number of parallel bandit problems.

\subsection{Organization} 
 
We begin by introducing the notations and problem formulation in \cref{sec:setting}. In \cref{sec:brittleness}, we present two families of examples that demonstrate how \LinTS{} without inflation can incur linear regret. Next, we propose our \TSAI{} algorithm and provide its theoretical analysis in \cref{sec:improved-ts,sec:constants}, and by empirical simulations in   \cref{sec:simulations}. Finally, we conclude in \cref{sec:discussion}, and relegate the proofs to the appendices.
\section{Setting and Notation}
\label{sec:setting}
\def\Matrix{\mathbf{M}}
\def\SingVals{\mathbf{D}}

We begin by introducing the notations that will be used throughout the paper. For any positive integer $n$, we denote the set ${1,2,\cdots,n}$ as $[n]$. Let $\SymCovMatrix$ be a positive semi-definite $n$ by $n$ matrix, and let $\Arm$ be any vector in $\IR^n$. We define the notation $\Norm{\Arm}_{\SymCovMatrix}$ for $\sqrt{\Arm^\top\SymCovMatrix \Arm}$. For a matrix $\Matrix$ with singular values $\sigma_1\geq\cdots\geq\sigma_n$, we define its operator norm as $\NormOp{\Matrix}:=\sigma_1$, and its trace norm (or nuclear norm) as $\NormNuc{\Matrix}:=\sum_{i\in[n]}\sigma_i$. To represent asymptotic upper and lower bounds, we use the standard notations $\Order{\cdot}$ and $\LowerOrder{\cdot}$, respectively. When logarithmic terms are suppressed, we use the notations $\OrderLog{\cdot}$ and $\LowerOrderLog{\cdot}$, and defer a formal definition to \cite{cormen2009introduction}. Lastly, we denote the cumulative distribution function (CDF) of the standard normal distribution by $\Phi(\cdot)$, and the $n$-dimensional identity matrix by $\II_n$.

Let $(\ArmSet_t)_{t=1}^T$ be a sequence of $T$ random compact subsets of $\IR^d$ where $T\in\IN$ is the time horizon. We further assume that $\Norm{\Arm}_2\leq\ArmBound$ for all $\Arm\in\ArmSet_t$ almost surely. A policy $\Policy$ sequentially interacts with the environment in $T$ rounds. At time $t\in[T]$, it receives action set $\ArmSet_t$ and chooses an action $\ChosenArm_t\in\ArmSet_t$ and receives a stochastic reward $\RewardValue_t=\Inner[\big]{\Param,\ChosenArm_t}+\eps_t$ where $\eps_t$ is the reward noise and 
 $\Param$ is an unknown (and potentially random) vector of parameters. By $\OptimalArm_t\in\ArmSet_t$ we denote the arm with maximum expected reward.
We denote the history of observations up to time $t$ by $\HistoryPlus_{t}$. More precisely, we define 
\[
\HistoryPlus_t:=(\ArmSet_1,\ChosenArm_1,\RewardValue_1,\cdots,\ArmSet_{t-1},\ChosenArm_{t-1},\RewardValue_{t-1},\ArmSet_t)\,.
\]
In this model, a \emph{policy} $\Policy$ is formally defined as a (stochastic) function that maps $\HistoryPlus_t$ to an element of $\ArmSet_t$.

We compare policies through their cumulative Bayesian regret defined as
\begin{align*}
    \Regret(T,\pi):=\sum_{t=1}^{T}\Expect*{\sup_{\Arm\in\ArmSet_t}\Inner[\big]{\Param,\Arm}-\Inner[\big]{\Param,\ChosenArm_t}}.
\end{align*}
Recall that  the expectation is taken with respect to the entire three sources of randomness in our model, including the prior distribution on $\Param$. The frequentist regret bounds also follow by taking the prior distribution to be the measure that puts all the mass on a single deterministic vector.
\section{Bayesian analyses are brittle}
\label{sec:brittleness}
In this section, we demonstrate that \LinTS{} may incur linear regret when the assumptions are \emph{slightly} violated. 
Our analysis reveals that when \LinTS{} employs an inaccurate prior or noise distribution, the Bayesian regret (and frequentist regret) can exhibit a linear growth rate. \footnote{This does not contradict the minimax optimal bound obtained by \cite{russo2014learning}. Their analysis assumes that \LinTS{} has access to the true prior and noise distributions, which is a stronger assumption than the one made here.}. To be more specific, we establish that a linear regret can occur,  when the dimensionality of the problem satisfies $d=\LowerOrder{\log T}$.  
We begin by offering an intuitive explanation of these examples in \cref{subsec:intuition}, after which we present the examples in \cref{subsec:example-changing-action-set,subsec:example-fixed-action-set}. The former employs action sets that vary over time, while the latter employs fixed action sets.


\subsection{Intuition}\label{subsec:intuition}

Here we construct a vanilla example where an adaptive adversary causes \LinTS{} to fail by adaptively choosing bad action sets. But, this example is chiefly intended to develop the intuition behind our main examples, where \emph{the action sets are selected independently from the history}. Rigorous proofs are provided for the examples in the next subsection. 

\begin{algorithm}[t]
	\caption{Linear Thompson sampling (\LinTS{})}
	\label{alg:proper-ts}
	\begin{algorithmic}[1]
		\REQUIRE Inflation parameter $\Inflation$.
		\STATE Initialize $\SymCovMatrix_1\gets\lambda\II$ and $\Estimator_1\gets0$
		\FOR{$t=1,2,\cdots$}
		\STATE Observe $\ArmSet_t$
		\STATE Sample $\TsSample_t\sim\Normal{\Estimator_t,\:\Inflation^2\SymCovMatrix_t}$
		\STATE $\ChosenArm_t\gets \Argmax_{\Arm\in\ArmSet_t}\Inner[\big]{\Arm,\TsSample_t}$
		\STATE Observe reward $Y_t$
		\STATE $\SymCovMatrix_{t+1}^{-1}\gets\SymCovMatrix_t^{-1}+\ChosenArm_t\ChosenArm_t^\top$
		\STATE $\Estimator_{t+1}\gets \SymCovMatrix_{t+1}\left(\SymCovMatrix_t^{-1}\Estimator_t+ \ChosenArm_t\RewardValue_t \right)$
		\ENDFOR
	\end{algorithmic}
\end{algorithm}

Next, somewhat counter-intuitively, we first state a positive result, a sufficient condition that leads to a sub-linear regret bound for \LinTS{}. Therefore, a counter-example in which \LinTS{} would fail must violate that sufficient condition which gives us intuition on how to construct counter-examples. We also note that,  this result is for the slightly more general version of \LinTS{} where the posterior distribution is inflated by a positive parameter $\Inflation$ (\cref{alg:proper-ts}). We will prove a more general version of this theorem in \cref{sec:improved-ts-proofs}.
\begin{thm}\label{thm:expected-regret-bound}
If \cref{alg:proper-ts} satisfies
\begin{align}
    \Prob*{
        \sup_{\Arm\in\ArmSet_t}\Inner{\Arm,\TsSample_t}
        \geq
        \sup_{\Arm\in\ArmSet_t}\Inner{\Arm,\Param}
        \Given
        \Param,\HistoryPlus_t
    }
    \geq
    \OptProb\,, \label{eq:cond-4-expected-regret-bound}
\end{align}
whenever $\Norm{\Estimator_t-\Param}_{\SymCovMatrix_t^{-1}}\leq\Radius$, we then have
\begin{align}
    \Regret(T,\PolicyTS)
    \leq
    \OrderLog*{\frac{\Radius\Inflation}{\OptProb}\sqrt{\Dim T}}.
    \label{eq:expected-regret-bound}
\end{align}
\end{thm}
Recall that the optimal action at time $t$ is denoted by $\OptimalArm_t$. We can now write
\begin{align*}
    \sup_{\Arm\in\ArmSet_t}\Inner{\Arm,\TsSample_t}
    -
    \sup_{\Arm\in\ArmSet_t}\Inner{\Arm,\Param}
    &\geq
    \Inner{\OptimalArm_t,\TsSample_t-\Param}.
\end{align*}
Therefore, a sufficient condition for \cref{eq:cond-4-expected-regret-bound} is that
\begin{align}\label{eq:suff-condition}
    \Prob*{
        \Inner{\OptimalArm_t,\TsSample_t-\Param}\geq0
        \Given
        \Param,\HistoryPlus_t
    }
    \geq
    \OptProb\,,
\end{align}
whenever $\Norm{\Estimator_t-\Param}_{\SymCovMatrix_t^{-1}}\leq\Radius$. Next notice that
\begin{align}\label{eq:decomposition}
    \Inner{\OptimalArm_t,\TsSample_t-\Param}
    &=
    \Inner{\OptimalArm_t,\TsSample_t-\Estimator_t}+
    \Inner{\OptimalArm_t,\Estimator_t-\Param}.
\end{align}
Looking at the above decomposition, we call $\Error_t:=\Param-\Estimator_t$ the \emph{error vector} and $\Compensator_t:=\TsSample_t-\Estimator_t$ the \emph{compensator vector}. The latter name is motivated by the observation that,
using \cref{eq:suff-condition} and \cref{eq:decomposition}, when
$
\Inner{\OptimalArm_t,\Compensator_t}
\geq
\Inner{\OptimalArm_t,\Error_t}
$ holds then
\begin{align*}
\sup_{\Arm\in\ArmSet_t}\Inner{\Arm,\TsSample_t}
\geq
\sup_{\Arm\in\ArmSet_t}\Inner{\Arm,\Param}
\end{align*}
holds as well. Thus, $\Compensator_t$ should \emph{compensate} for the underestimation of $\Inner{\OptimalArm_t,\Param}$ caused by $\Error_t$. While this inequality is only a \emph{sufficient} condition to obtain \cref{eq:cond-4-expected-regret-bound}, we demonstrate how it can be used to deceive \LinTS{}.

An adversary that knows $\Estimator_t$ and $\Param$ (thereby, $\Error_t$) can exploit \LinTS{} by showing an action set of the form $\{0,\Arm\}$ where $\Arm$ satisfies
\[
\frac{1}{2}\Inner{\Arm,\Error_t}\approx-\Inner{\Arm,\Estimator_t}\approx\Inner{\Arm,\Param}>0=\Inner{0,\Param}\,.
\] 
Therefore, $\Arm$ would be the optimal action (i.e., $A=\OptimalArm_t$).
In this case, \LinTS{} would choose $\Arm$ if and only if 
$\Inner{\Arm,\TsSample_t}>0$, which would then be approximately equivalent to
\[
\Inner{\Arm,\Compensator_t}
\geq
\frac12\Inner{\Arm,\Error_t}\,.
\]
But $A$ can be chosen to be align with $\Inner{\Arm,\Error_t}$ which would make it much larger than $\Inner{\Arm,\Compensator_t}$ because 
$\Compensator_t$ is an independent random vector, conditioned on the history. 
This would allow $\Arm$ to be chosen so that $\Inner{\Arm,\Error_t}\approx\Norm{\Arm}_2\Norm{\Error_t}_2$, whereas $\Inner{\Arm,\Compensator_t}\approx\frac1{\sqrt d}\Norm{\Arm}_2\Norm{\Compensator_t}_2$ with high probability. Therefore, \LinTS{} will select $\ChosenArm_t=0$ with a high probability while it is not the optimal arm. Moreover, $\ChosenArm_t=0$ reveals no more information about the true parameters. Hence, even if the same action set is shown in the next rounds, \LinTS{} will fail to detect the optimal arm. In the remaining, we will make this intuition more formal.

\subsection{Example 1: Noise reduction and changing action sets}\label{subsec:example-changing-action-set}

In the previous section, we discussed how aligning the arm selection with the error vector can impact the performance. However, when the distributions used to compute the posterior distribution in \LinTS{} do not match the actual distributions, a marginal bias can occur in the error vector $\Error_t$. This can have a negative impact on the performance of \LinTS{} when the action set is appropriately chosen. Remarkably, this bias can even occur when the data quality is improved by reducing the noise variance. Importantly, the action sets can be constructed ahead of time without knowledge of the algorithm decisions. In this subsection, we describe our strategy for proving these results. First, we construct small problem instances in which $\TsSample_t$ is \emph{marginally biased}. We then demonstrate that by combining independent copies of these biased instances, \LinTS{} can incur linear Bayesian regret.

\begin{rem}\label{rem:example-1-guide}
We study \LinTS{} as shown in \cref{alg:proper-ts} with $\Inflation=1$ and $\lambda=1$.  This means we study  \LinTS{} that does not inflate posterior variance and assumes the noise variance and prior variance are equal to $1$.  In the example that will be constructed below, the true noise and prior variance will be $\tau^2$ and $\sigma^2$, respectively. Then, the main result of the section will show that when $\tau\neq\sigma$, \LinTS{} achieves linear regret. But note that $\tau\neq\sigma$ means at least one of $\tau$ or $\sigma$ is not equal to $1$, which means at least one of noise or prior variance does not match the one that \LinTS{} assumes.
\end{rem}

\paragraph{Bias-introducing action sets.}
In this section, we construct an example in which $\TsSample_t$ is marginally biased, provided that either the prior distribution or the noise distribution mismatches the one that \LinTS{} uses. 
Fix $\sigma^2,\tau^2\geq 0$ and let $\Param\sim\Normal{0,\sigma^2\II_2}$ be the vector of unobserved parameters. At time $t\in\{1,2,3\}$, we reveal the following action sets to the policy:
\begin{align*}
    \ArmSet_t:=
    \begin{cases}
        \{e_1\} & \text{if $t=1$,}\\
        \{e_2\} & \text{if $t=2$,}\\
        \{e_1,e_2\} & \text{if $t=3$}\,,
    \end{cases}
\end{align*}
where $e_1$ and $e_2$ are standard basis vectors in $\IR^2$. For $t\leq2$, \LinTS{} has only one choice $e_t$ and thus $\ChosenArm_t=e_t$. Assume that reward $Y_t=\Param_t+\eps_t$ is revealed to the algorithm where $\eps_t\sim\Normal{0,\tau^2}$. At time $t=3$ for the first time, \LinTS{} has two choices. Let $i$ in $[2]$ be such that $\ChosenArm_3=e_i$. Then, $Y_3=\Param_i+\eps_3$ is provided to the algorithm where $\eps_3\sim\Normal{0,\tau^2}$. The following key lemma proves that $\Estimator_4$ is marginally biased when $\tau\neq\sigma$.
\begin{lem}
\label{lem:bias-quant}
Let $V=e_1+e_2$. For any $\sigma,\tau\geq0$, we have
\begin{align}
    \Inner[\big]{V,\Expect[\big]{\Estimator_4}}
    &=
    \frac{\left(\sigma^2-\tau^2\right)\BiasCoeff}{6\sqrt{\sigma^2+\tau^2+2}}\,,
    \label{eq:v-bias-closed-form}
\end{align}
where $\BiasCoeff:=\Expect*{\max\{A, B\}}>0$ with $A$ and $B$ being two independent standard normal random variables. Furthermore, for a positive constant $C$, $\Estimator_4$ satisfies
\begin{align}
    \Expect*{\exp\left(s\Inner[\big]{V,\Estimator_4-\Expect[\big]{\Estimator_4}}\right)}
    \leq
    \exp\left[Cs^2 (\sigma+\tau+\sqrt{2})^2\right]\,,
    \label{eq:bias-mgf}
\end{align}
for all $s\in\IR$.
\end{lem}
This finding illuminates that a bias emerges in the posterior mean estimate yielded by the \LinTS{} algorithm due to a combination of two factors: \emph{distribution mismatch} and \emph{\randombias{} bias}. The former refers to the discrepancy between \LinTS's assumption on the true prior variance and the true noise variance, which is shown in \cref{rem:example-1-guide} to be equivalent to $\sigma^2-\tau^2\neq 0$. The latter stems from the randomization inherent in \LinTS{}, resulting in the introduction of a positive term $\beta$.

We present a proof of \cref{lem:bias-quant} in \cref{sec:brittleness-proofs}. However, we shall here provide a brief sketch of the proof. Firstly, we demonstrate that $ \Inner{V,\Estimator_4}$ can be expressed as a linear combination of $ \Inner{V,\Estimator_3}$, $\Param_i-\eps_i$, and $\eps_3$. The first and third terms are unbiased, with a mean of zero, thus our attention is focused on $\Expect{\Param_i-\eps_i}$. Subsequently, we establish that $\Expect{\Param_i}$ and $\Expect{\eps_i}$ are equal to a shared constant, multiplied by $\sigma^2\,\Expect{\TsSample_{3,i}}$ and $\tau^2\,\Expect{\TsSample_{3,i}}$, respectively. 
Because the expected value of $\TsSample_{3,i}$ is proportional to  $\beta$, it is now evident to trace back the origin of both $(\sigma^2-\tau^2)$ and $\beta$ in \cref{eq:v-bias-closed-form} to the distribution mismatch and \randombias{} bias.

\paragraph{Stacking biased blocks.}
By combining independent copies of the above example, we prove that \LinTS{} can choose an incorrect action for at least $\exp(\LowerOrder{d})$ rounds. Let $d$ be a positive integer and define $\Param\sim\Prior=\Normal{0,\sigma^2\II_{2d}}$. We will construct a $2d$-dimensional linear bandit setting where in the first $3d$ rounds, the action sets of the type introduced above for each pairs
$(\Param_{2i-1},\Param_{2i})$ for $i\in[d]$ are presented to the algorithm. Namely, define
\begin{align}
    \begin{split}
    \ArmSet_t
    :=
    \begin{cases}
        \{e_t\} & \text{if $t\leq 2d$,}\\
        \{e_{2(t-2d)-1},e_{2(t-2d)}\} & \text{if $2d+1\leq t\leq 3d$,}\\
        \{0, \Arm\} & \text{otherwise,}
    \end{cases}
    \end{split}
    \label{eq:stacked-arm-sets}
\end{align}
where
$
    \Arm
    :=
    ({\Sign{\tau^2-\sigma^2}}/{\sqrt{d}})\cdot\sum_{i=1}^{2d}e_i$.
Note that, due to the term $\Sign{\tau^2-\sigma^2}$, $\Arm$ is in the opposite direction of the marginal bias of $\Estimator_{3d+1}$. Therefore, \LinTS{} will be less likely to select $\Arm$, and this will be an incorrect decision. Formally,
the following key lemma, proved in \cref{sec:brittleness-proofs}, states that with constant probability, $\Arm$ is the optimal action, while \LinTS{} \emph{perceives} it as suboptimal, with an enormous gap.
\begin{lem}
    \label{lem:misperception}
    For positive constants $p_0=\frac{1}{2}(1-\Phi(1))$ and $C_1(\sigma,\tau):=\frac{|\sigma^2-\tau^2|\BiasCoeff}{6\sqrt{\sigma^2+\tau^2+2}}$, the following holds
    \begin{align*}
        \Prob*{\Inner[\big]{\Param,\Arm}\geq\sqrt 2\sigma~~\text{and}~~\Inner[\big]{\Estimator_{3d+1},\Arm}\leq-\frac{C_1(\sigma,\tau)\sqrt d}{2}}\geq p_0\,.
    \end{align*}
\end{lem}
We denote the event in the above Lemma by $\BadEvent$. Conditional on this event, for all $t>3d$, the optimal arm is $\Arm$, and the regret incurred by choosing the action $0$ is at least $\sqrt 2\sigma$. Moreover, let $q$ be the probability of choosing $\Arm$ at $t=3d+1$. As we will see, when $\tau\neq \sigma$, this probability is exponentially small as a function of $d$.  The probability of selecting $\Arm$ in the next round remains unchanged, whenever $\Arm$ is not chosen. This observation holds up to the first time that $\Arm$ is picked, which can, in turn, take an exponentially long time. By making this argument rigorous, we can state the following proposition which is proved in \cref{sec:brittleness-proofs}.
\begin{prop}
\label{prop:noise-reduction-failure}
In Example 1, when $\sigma\neq\tau$ and $T\leq\exp(\LowerOrder{d})$, we have
$
    \Regret(T,\PolicyTS)
    \geq
    \LowerOrder{T}\,.$%
\end{prop}
An immediately corollary of this result is that when $d$ is comparable to $\log T$, which can naturally occur in practice, \LinTS{} incurs a linear regret.
\begin{cor}
	\label{cor:noise-reduction-failure}
	In Example 1, when $d=\LowerOrder{\log T}$ and $\sigma\neq\tau$,  we have
	\begin{align*}
		\Regret(T,\PolicyTS)
		\geq
		\LowerOrder{T}\,.
	\end{align*}
\end{cor}
%
%
Drawing upon \cref{rem:example-1-guide}, the aforementioned result underscores that a discrepancy between the actual noise or prior distribution and those which \LinTS{} presumes, leads to linear regret. Interestingly, a specific instance of this circumstance arises when $\tau=0$, thereby highlighting a scenario in which \LinTS{} falters despite being provided with superior-quality data than it assumes.

\subsection{Example 2: Mean shift and fixed action sets}\label{subsec:example-fixed-action-set}

In this subsection, we construct an example in which \LinTS{} incurs linear Bayes regret while \emph{the action set is fixed over time}. Like the previous example, we assume \LinTS{}  does not inflate posterior variance and assumes the noise and prior distribution are both standard normal.
Let $\mu,\sigma,\tau>0$ be fixed, and for $d\in\IN$, set the prior distribution to be $\Prior:=\Normal{\mu\One_{3d},\sigma^2\II_{3d}}$. We now reveal the action set $\ArmSet_t:=\{0, \Arm', \Arm\}$ to \LinTS{} for all $t\in[T]$ where
\begin{align}
\begin{split}
    \Arm'&:=-\frac{1}{\sqrt d}\sum_{i=1}^{d}e_i,\\
    \Arm&:=\frac{1}{\sqrt d}\sum_{i=d+1}^{3d}e_i-\frac{1}{\sqrt d}\sum_{i=1}^{d}e_i\,.
\end{split}
\label{eq:fixed-arm-set}
\end{align}
The next proposition, proved in \cref{sec:brittleness-proofs}, highlights the key observations about why \LinTS{} fails in this simple setting.
\begin{prop}
\label{prop:mean-shift-failure}
For fixed $\mu,\sigma>0$, and for sufficiently large $d$, we have
\begin{enumerate}
\item $\Inner{\Param,\Arm'}\leq-\frac12\mu\sqrt d\leq\frac12\mu\sqrt d\leq\Inner{\Param,\Arm}$, with probability at least $\frac78$.

\item $\ChosenArm_1=\Arm'$ with probability $\frac14$.

\item Conditional on $\ChosenArm_1=\Arm'$,  $\Inner[\big]{\Estimator_2,A}$ and $\Inner[\big]{\Estimator_2,A'}$ are less than $-\frac18\mu\sqrt d$, with probability at least $\frac{15}{16}$.

\item Conditional on $\ChosenArm_1=\Arm'$, $\ChosenArm_2\neq0$ with probability at most $\exp(-\LowerOrder{d})$.

\item For all $T\leq\exp(\LowerOrder{d})$, $\Regret(T,\PolicyTS)\geq\LowerOrder{T\sqrt d}$.
\end{enumerate}
\end{prop}
\begin{rem}\label{rem:corr-failure}
One can slightly modify the proof to obtain a similar result for
$
    \Prior
    :=
    \Normal{0,\sigma^2\II_{3d}+\rho\One_{3d}\One_{3d}^\top}
$.
It is easy to see that for any arbitrary constant $\rho$, the same rate as in \cref{eq:mean-shift-lower-bound} is achievable. Also, for $\rho=d^{-\alpha}$ where $\alpha < 1$, one can still get non-trivial results.
\end{rem}
An immediate implication of \cref{prop:mean-shift-failure} and \cref{rem:corr-failure} is that if there exists a mismatch between \LinTS{} and the true prior at the mean or variance level in Example 2 with a fixed action set, then \LinTS{} incurs linear regret for an extended exponential period. Similar to Example 1, we obtain the following corollary.
\begin{cor}
	\label{cor:mean-shift-failure}
	In Example 2, when $d=\LowerOrder{\log T}$ and the true prior
 $\Prior$ is either $\Normal{\mu\One_{3d},\sigma^2\II_{3d}}$ or  $\Prior=\Normal{0,\sigma^2\II_{3d}+\rho\One_{3d}\One_{3d}^\top}$, 
 we have
	\begin{align*}
		\Regret(T,\PolicyTS)
		\geq
		\LowerOrder{T}\,.
	\end{align*}
\end{cor}
In summary, we have presented a simple setting in which \LinTS{} incurs linear Bayes regret even when the action set is fixed over time.

\begin{rem}\label{rem:adversary-knowledge}
	It is worth noting that the design of the action sets in Example 1 (or Example 2) necessitates solely the understanding of the sign of $\tau^2-\sigma^2$ (or sign of $\mu$). In the context of a competition between two competiting platforms $\mathcal{P}_1$ and $\mathcal{P}_2$ from \cref{sec:intro}, $\mathcal{P}_1$ only needs to be aware of the direction of inconsistency between the true reward distribution and the one which is presumed by $\mathcal{P}_2$.
\end{rem}

\section{Thompson Sampling with Adaptive Inflation}
\label{sec:improved-ts}

In this section, we present an alternative approach to improve the inflation parameter in \LinTS{} and enhance its performance, subject to additional assumptions.
To facilitate a better understanding of our proposed method, we first provide the intuition behind the development of the conditions. These intuitions follow the discussion in \cref{subsec:intuition}, with a focus on exploring the mechanisms that enable \LinTS{} to succeed, rather than those that may cause it to fail. Building on these insights, we introduce Thompson Sampling with Adaptive Inflation (\TSAI{}), an algorithm that adaptively adjusts its inflation parameter to meet the aforementioned conditions. Finally, we state our informal regret bound for \TSAI{} and defer its formal proof to \cref{sec:improved-ts-proofs}.

Note that at any time in \LinTS{}, the posterior mean $\Estimator_t$ is the ridge estimator for the parameter $\Param$, given the actions and their observed rewards in prior rounds. Additionally, we define $\ConfSet_t$ as the confidence set centered around $\Estimator_t$, which is constructed as part of the OFUL algorithm and contains $\TsSample_t$ and $\Param$ with high probability. More information on the construction and properties of these confidence sets can be found in \citep{abbasi2011improved}.

Assume that $d=\LowerOrder{\log T}$.  As in \cref{subsec:intuition}, and for the sake of building intuition, we first restrict our attention to action sets of the form $\{\OptimalArm_t, 0\}$ where $\OptimalArm_t$ is the optimal arm, i.e., $\Inner*{\Param,\OptimalArm_t}>0$. \LinTS{} chooses $\OptimalArm_t$ only if
\begin{align}
\Inner[\big]{\TsSample_t,\OptimalArm_t}>0\,.
    \label{eq:LinTS-condition}
\end{align}

\begin{figure}[th]
\centering
\subcaptionbox{{\footnotesize Actual confidence set}}{
    \includegraphics[width=0.4\linewidth]{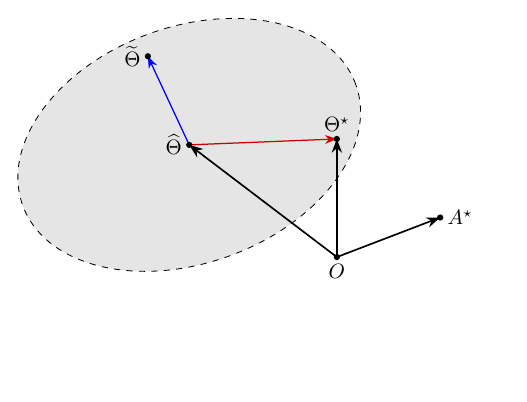}
}
\subcaptionbox{{\footnotesize Translated confidence set}}{
    \includegraphics[width=0.3\linewidth]{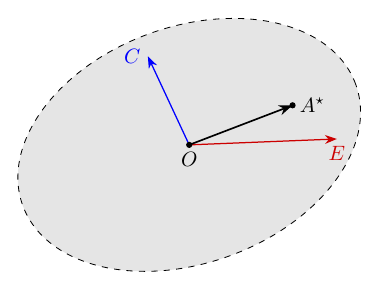}
}
\vspace{3mm}
\caption{A typical setting for $\OptimalArm_t$, $\Param$, $\Estimator_t$, and $\TsSample_t$. The compensation vector $C=\TsSample_t-\Estimator_t$ and the error vector $E=\Param-\Estimator_t$ are defined in \cref{subsec:intuition}, and $O$ denotes the origin.}
\label{fig:optimism-intro}
\end{figure}

\paragraph{Compensation inequality.} 
By decomposing the left-hand side of \cref{eq:LinTS-condition} as we did in \cref{eq:decomposition},  a sufficient condition for \cref{eq:LinTS-condition} to hold is
\begin{align}
    \Inner[\big]{\TsSample_t-\Estimator_t,\OptimalArm_t}
    \geq
    \Inner[\big]{\Param-\Estimator_t,\OptimalArm_t}\,,
    \label{eq:compensation}
\end{align}
which is also equivalent to
$\Inner{C,\OptimalArm_t}\geq\Inner{E,\OptimalArm_t}$
with $C$ and $E$ defined in \cref{subsec:intuition} and illustrated in \cref{fig:optimism-intro}.

\begin{figure}[th]
	\centering
	\includegraphics[scale=.8]{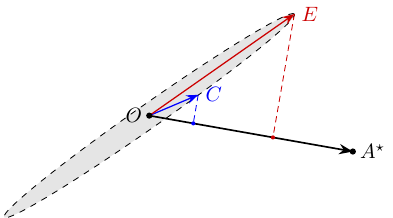}
	\caption{A thin confidence set.}
	\label{fig:optimism-thinness}
\end{figure}

OFUL explicitly seeks $\TsSample_t\in\ConfSet_t$ that maximizes the left-hand side of \cref{eq:compensation}, and as $\Param\in\ConfSet_t$ with high probability, the desired ``compensation inequality'' holds, and $\OptimalArm_t$ is selected with high probability. \LinTS{}, on the other hand, follows a stochastic approach and resorts to a randomly sampled $\TsSample_t$, that with high probability resides in $\ConfSet_t$, to solve \cref{eq:compensation}. Since $\Estimator_t$ is the ridge estimator for the collected data thus far, in a fixed design setting (which is not true in our bandit problem with adaptively collected data) the error vector $E$ will be pointing in a random direction. Therefore, provided that $\OptimalArm_t$ is independent of $E$, we have
\begin{align}
    \Abs[\big]{\Inner[\big]{E,\OptimalArm_t}}
    \approx
    \Order*{\frac1{\sqrt\Dim}\Norm{E}_2\cdot\Norm{\OptimalArm_t}_2}.
    \label{eq:easy-error}
\end{align}
The same expression also holds for $\Abs[\big]{\Inner[\big]{C,\OptimalArm_t}}$; therefore, the compensation inequality holds with constant probability. To summarize our observation, \cref{eq:easy-error} holds  if the error vector $E$ is distributed in a random direction that is independent of the optimal action $\OptimalArm_t$.
The crucial point in the analysis of \LinTS{} in the Bayesian setting is that the error vector is in a random direction whenever \LinTS{} has access to the true prior and noise distribution. In \cref{sec:brittleness}, nonetheless, we have shown that this condition is violated if \LinTS{} uses an incorrect prior or noise distribution in computing the posterior. \cite{agrawal2013thompson,abeille2017linear} take a conservative approach and propose to inflate the posterior distribution by a factor of $\OrderLog{\sqrt{d}}$ that inflates $C$ by the same factor
to ensure $\Inner*{C,\OptimalArm_t}\geq\Inner*{E,\OptimalArm_t}$ holds with constant probability. \emph{We present an alternative approach that leverages the randomness of the optimal action to reduce the need for exploration}. The following assumption requires the \emph{optimal arm} (rather than the error vector) to be distributed in a random direction.
\begin{asmp}
\label{as:optimal-diversity}
\def\Vec{V}
Assuming that for any $\Vec\in\IR^{d}$ with $\Norm{\Vec}_{2}=1$, the inequality
\begin{align}
	\Inner*{\OptimalArm_t,\Vec}
	\leq
	\frac{\DivParam}{\sqrt{d}}\Norm{\OptimalArm_t}_2
	\label{eq:diversity}
\end{align}
holds with a probability of at least $1-\frac1{T^2}$, where $\DivParam$ is a fixed parameter.
\end{asmp}
Unfortunately, this condition alone does not suffice to reduce the inflation rate of the posterior distribution. To see this, consider a case in which the largest eigenvalue of $\SymCovMatrix_t$ is much larger than the other eigenvalues of $\SymCovMatrix_t$; thereby, $\NormOp{\SymCovMatrix_t}\approx\NormNuc{\SymCovMatrix_t}$ and $E$ points to the longest direction of the confidence set. \cref{fig:optimism-thinness} illustrates this situation. In this case, we have
\begin{align*}
    \Abs[\big]{\Inner[\big]{E,\OptimalArm_t}}
    &\approx
    \frac{\Norm[\big]{E}_2\cdot\Norm{\OptimalArm_t}_2}{\sqrt d}\\
    &\approx
    \frac{\sqrt{d\,\NormOp{\SymCovMatrix_t}}\cdot\Norm{\OptimalArm_t}_2}{\sqrt d}\\
    &=
    \sqrt{\NormOp{\SymCovMatrix_t}}\cdot\Norm{\OptimalArm_t}_2\,.
\end{align*}
The second approximation utilizes two observations. First, it exploits the fact that $\Norm{E}_2\approx \Norm{E}_{\SymCovMatrix_t^{-1}}\cdot \Norm{E}_{\SymCovMatrix_t}$, which is justified by the alignment of $E$ with the longest direction of $\SymCovMatrix_t$. Second, it relies on the high probability event that $E$ is contained in $\ConfSet_t$, implying that $\Norm{E}_{\SymCovMatrix_t^{-1}}$ is of the order $\sqrt{d}$.

However, it follows from the definition of \LinTS{} that $\Inner[\big]{\TsSample_t-\Estimator_t,\OptimalArm_t}\sim\Normal[\big]{0,\Inflation^2\Norm{\OptimalArm_t}_{\SymCovMatrix_t}^2}$. Assuming that $\Expect{\OptimalArm_t\OptimalArm_t{}^\top}\approx\Eye_d$, we realize that $\Expect[\big]{\Norm{\OptimalArm_t}_{\SymCovMatrix_t}^2}\approx\NormNuc{\SymCovMatrix_t}$. This suggests $\Norm{\OptimalArm_t}_{\SymCovMatrix_t}$ is proportional to 
\[
\sqrt{\NormNuc{\SymCovMatrix_t}/\Dim~}\cdot\Norm{\OptimalArm_t}_2\,.
\]
Now, we can see that \cref{as:optimal-diversity} is not sufficient for ensuring \cref{eq:compensation} as we have
\begin{align*}
    \Abs[\big]{\Inner[\big]{E,\OptimalArm_t}}
    \approx
    \sqrt{\NormOp{\SymCovMatrix_t}}\cdot\Norm{\OptimalArm_t}_2
    \gg
    \sqrt{\frac{\NormNuc{\SymCovMatrix_t}}d}\cdot\Norm{\OptimalArm_t}_2
    \approx
    \Abs[\big]{\Inner[\big]{C,\OptimalArm_t}}.
\end{align*}
This observation implies the necessity of the inflation rate of $\OrderLog{\sqrt d}$ when the eigenvalues of $\SymCovMatrix_t$ differ in magnitude significantly. To make this notion precise, we define the \emph{thinness coefficient} of a positive definite matrix $\SymCovMatrix$ to be
\begin{align*}
    \Thinness(\SymCovMatrix)
    :=
    \sqrt{\frac
        {d\cdot\NormOp{\SymCovMatrix}}
        {\NormNuc{\SymCovMatrix}}
    }.
\end{align*}
The following assumption requires $\OptimalArm_t$ to be distributed in a way that benefits from low thinness.
\begin{asmp}
\label{as:optimal-random-friendly}
For $\ThinnessBound,\ExploreRatio>0$, we have
\begin{align}
\Norm{\OptimalArm_t}_{\SymCovMatrix}
    \geq
    \ExploreRatio\sqrt{\frac{\NormNuc{\SymCovMatrix}}\Dim}\cdot\Norm{\OptimalArm_t}_2\,,
    \label{eq:random-friendly}
\end{align}
with probability at least $1-\frac1{T^2}$, for any positive semi-definite matrix $\SymCovMatrix$ with $\Thinness(\SymCovMatrix)\leq\ThinnessBound$.
\end{asmp}
\cref{as:optimal-diversity,as:optimal-random-friendly} are sufficient for reducing the inflation parameter, whenever $\Thinness(\SymCovMatrix_t)\leq\ThinnessBound$. In the following theorem, we state our regret bound informally. The formal version of this result (\cref{thm:lin-ts-regret}) and its proof can be found in \cref{sec:improved-ts-proofs}.
\begin{thm}[Informal]
\label{thm:lin-ts-regret-informal}
Assume that $\Inner{\Arm,\Param}\in[-1,1]$ for all $\Arm\in\ArmSet_t$ almost surely. Then, under \cref{as:optimal-diversity,as:optimal-random-friendly}, we have
\begin{align*}
    \sum_{t=1}^T
    \Inner{\OptimalArm_t,\Param}-\Inner{\ChosenArm_t,\Param}
    &\leq
    \OrderLog{\Inflation\Dim\sqrt{T}}
    +
    2
    \sum_{t=1}^{T}\Indicator(\Thinness(\SymCovMatrix_t)>\ThinnessBound)\,,
\end{align*}
with probability at least $1-\frac3T$, provided that the inflation prameter $\Inflation$ of \LinTS{} satisfies $\Inflation\geq\frac{\DivParam\ThinnessBound}{\ExploreRatio}\cdot\OrderLog{1}$.
\end{thm}
In \cref{sec:simulations}, for a concrete example from \citep{russo2014learning}, we empirically show that $\Thinness(\SymCovMatrix_t)<\ThinnessBound$ holds for a small value of $\ThinnessBound$ with high probability in our simulations. However, this condition is \emph{not} a mere property of the environment and depends on the interactions of \LinTS{} with the environment. Nonetheless, notice that $\Thinness(\SymCovMatrix_t)$ is observable, and the policy can intervene if $\Thinness(\SymCovMatrix_t)>\ThinnessBound$ for many rounds. This is the main idea behind our \TSAI{} algorithm presented in \cref{alg:tsai}. 
Note that the parameter $\Radius_t$, formally defined in \cref{eq:Radius}, 
plays the role of the factor $\sqrt{d}$ inflation as in \citep{agrawal2013thompson} and \citep{abeille2017linear}. 
However, in \TSAI{}, such inflation is only performed when the adaptively calculated thinness parameter $\Thinness(\SymCovMatrix_t)$ is too large. Otherwise, a constant inflation parameter $\Inflation$ is used.

For the same example from \citep{russo2014learning}, in \cref{sec:constants}, we will theoretically prove that \cref{as:optimal-diversity,as:optimal-random-friendly} hold.

\begin{rem}[OFUL with smaller confidence intervals]\label{rem:oful-smaller-intervals}
	Our proof in \cref{sec:improved-ts-proofs} reveals that, under  \cref{as:optimal-diversity,as:optimal-random-friendly}, it is possible to improve the performance of the OFUL algorithm by running it with smaller confidence sets that are reduced by a factor of order $\sqrt{d}$. It is worth noting that while the reduced confidence sets only impact the constants in the regret bound, but they may lead to improved empirical performance of the OFUL algorithm.
\end{rem}

\begin{rem}[Towards relaxing \cref{as:optimal-diversity,as:optimal-random-friendly}]\label{rem:general-assumptions}
It is noteworthy that the \cref{as:optimal-diversity,as:optimal-random-friendly} primarily serve to facilitate the theoretical analysis. 
From the proof of \cref{lem:lin-ts-optimism}, 
one requires the inflation parameter $\Inflation$ to satisfy 	
\begin{align}
	\Inflation
	\ge 
	\sup_{V\in\IS^{\Dim-1}}\frac{\Inner*{\OptimalArm_t,V}\Radius\sqrt{\NormOp{\SymCovMatrix_t}}}{\Norm{\OptimalArm_t}_{\SymCovMatrix_t}}\,,
	\label{eq:data-driven-inflation}
\end{align}
with a probability of at least $1-T^{-2}$, where $\IS^{\Dim-1}$ denotes the $\Dim$-dimensional unit sphere. 
Hence, for problems where \cref{as:optimal-diversity,as:optimal-random-friendly} may not hold, a data-driven approach to setting the inflation parameter is to follow \cref{eq:data-driven-inflation}, provided that the structure of the problem allows to bound the supremum on the right-hand side of \cref{eq:data-driven-inflation}.
\end{rem}	

\begin{algorithm}[t]
	\caption{Thompson Sampling with Adaptive Inflation (\TSAI{})}
	\label{alg:tsai}
	\begin{algorithmic}[1]
		\REQUIRE Inflation parameter $\Inflation$ and thinness threshold $\ThinnessBound$.
		\STATE Initialize $\SymCovMatrix_1\gets\lambda\II$ and $\Estimator_1\gets0$
		\FOR{$t=1,2,\cdots$}
		\STATE Observe $\ArmSet_t$
		\IF{$\Thinness(\SymCovMatrix_t)>\ThinnessBound$}
	    \STATE  Sample $\TsSample_t\sim\Normal{\Estimator_t,\:\Radius_t^2\,\SymCovMatrix_t}$ where $\Radius_t$ is defined in \cref{eq:Radius}
	    \ELSE 
	    \STATE Sample $\TsSample_t\sim\Normal{\Estimator_t,\:\Inflation^2\SymCovMatrix_t}$
	    \ENDIF
		\STATE $\ChosenArm_t\gets \Argmax_{\Arm\in\ArmSet_t}\Inner[\big]{\Arm,\TsSample_t}$
		\STATE Observe reward $Y_t$
		\STATE $\SymCovMatrix_{t+1}^{-1}\gets\SymCovMatrix_t^{-1}+\ChosenArm_t\ChosenArm_t^\top$
		\STATE $\Estimator_{t+1}\gets \SymCovMatrix_{t+1}\left(\SymCovMatrix_t^{-1}\Estimator_t+ \ChosenArm_tY_t \right)$
		\ENDFOR
	\end{algorithmic}
\end{algorithm}

\section{Simulations}
\label{sec:simulations}

In this section, we first, in \cref{subsec:avg-failure}, provide a numerical validation for the examples in \cref{sec:brittleness} that demonstrate two scenarios under which LinTS fails to choose the best action for an exponentially long time horizon. Then, in \cref{sec:simulations-thinness}, we compare the performance of our \TSAI{} with Bayesian and frequentist \LinTS{} in different settings.

\subsection{Average failure time of \LinTS{}} \label{subsec:avg-failure}
We provide two sets of simulations to validate the theoretical predictions of the two examples in \cref{sec:brittleness}.

\begin{figure}[th]
	\begin{center}
		\begin{subfigure}{0.45\textwidth}
			\includegraphics[width=\textwidth]{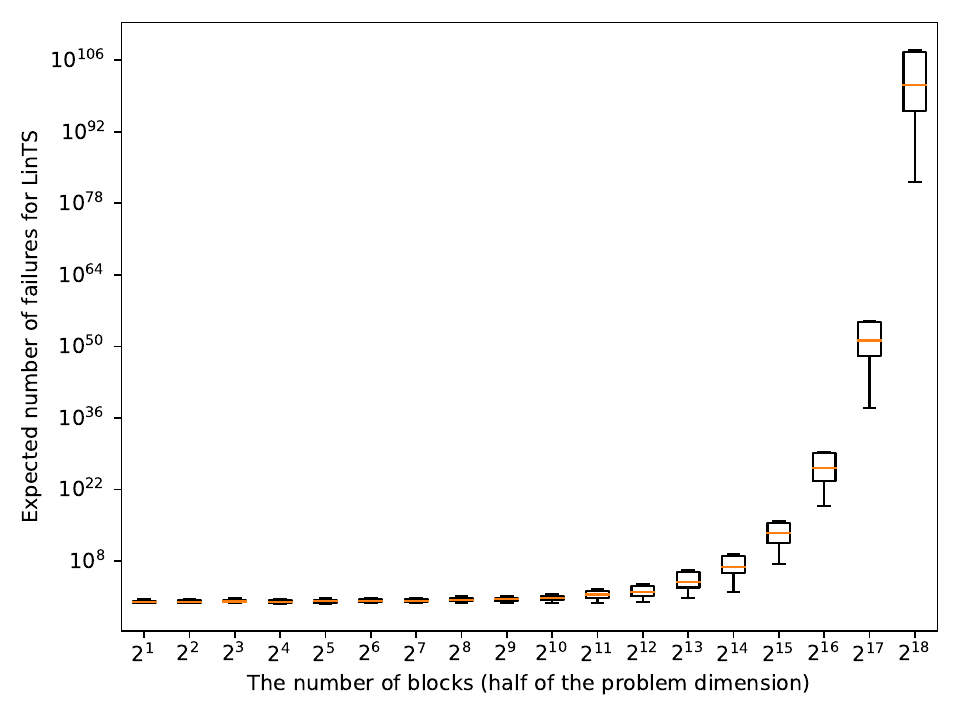}
			\caption{{\footnotesize Example 1}}
		\end{subfigure}
		\hfill
		\begin{subfigure}{0.45\textwidth}
			\includegraphics[width=\textwidth]{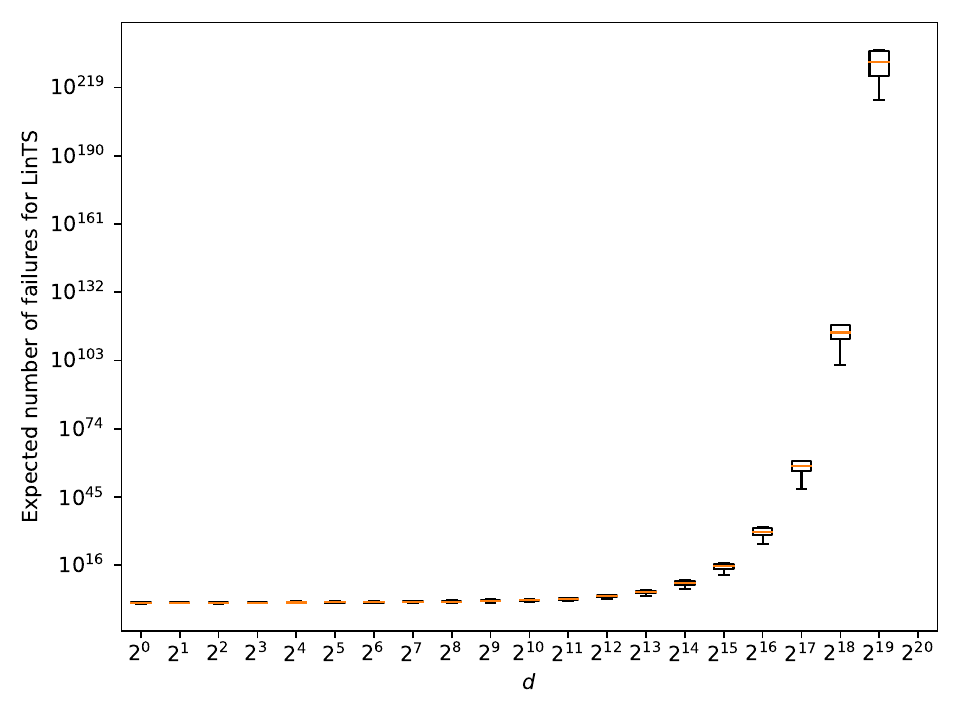}
			\caption{{\footnotesize Example 2 with $\mu=0.1$ and varying $d$}}
		\end{subfigure}
		\hfill
		\begin{subfigure}{0.45\textwidth}
			\includegraphics[width=\textwidth]{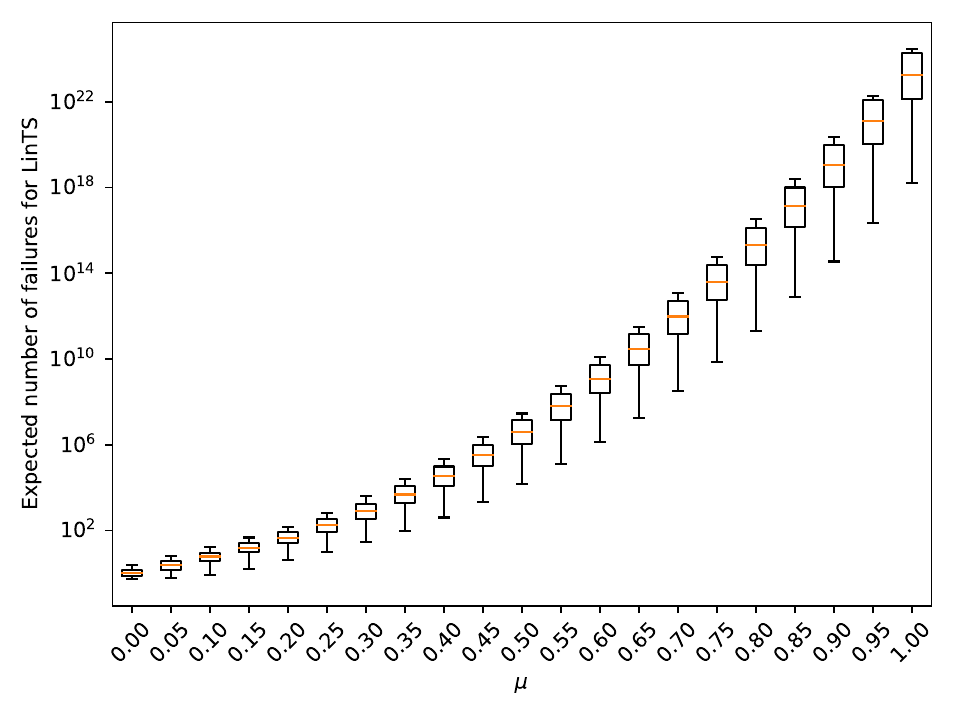}
			\caption{{\footnotesize Example 2 with $d=1000$ and varying $\mu$}}
		\end{subfigure}
		\vspace{3mm}
		\caption{Boxplots of $1/{p_i}$'s in Examples 1 and 2.}\label{fig:failures}
	\end{center}
\end{figure}

\paragraph{Noise reduction example.} In this simulation, for each $d\in\{2, 2^2, 2^3,\cdots,2^{18}\}$, we generate $\Param\sim\Normal{0,\Eye_{2d}}$ and execute \LinTS{} for $3d$ rounds using the action sets in \cref{eq:stacked-arm-sets}. The reward for choosing an action $\ChosenArm_t\in\ArmSet_t$ is simply given by $Y_t=\Inner[]{\Param,\ChosenArm_t}$. Therefore, no noise is added to the reward (i.e., $\tau=0$). After executing \LinTS{} for $3d$ rounds and obtaining $\Estimator_{3d+1}$, we compute the probability that $\Inner[]{\Estimator_{3d+1},\Arm}>0$.
Note that we can calculate this probability given that $\Estimator_{3d+1}$ is Gaussian and $\Arm$ is a multiple of $\One_{2d}$. 
Also, recall from \cref{subsec:example-changing-action-set} that, the complement of this event is when \LinTS{} incorrectly chooses action $0$, and under that scenario, the probability of selecting $\Arm$ in the next round stays the same. Hence, \LinTS{} would be expected to chose $0$ for $1/p$ time periods. Since $0$ is suboptimal with probability $1/2$, this means \LinTS{} would be expected to choose the suboptimal arm $1/(2p)$ time periods.
We repeat this procedure 100 times to obtain 100 values for $p$, denoted by $(p_i)_{i=1}^{100}$, and present a boxplot for the values $(1/(2p_i))_{i=1}^{100}$ in  \cref{fig:failures}(a), indicating the expected number of failures of \LinTS{} versus $d$. It is evident that, as predicted in 
\cref{subsec:example-changing-action-set},
 \LinTS{} selects the suboptimal action for at least $\exp(\LowerOrder{d})$ rounds.
  
\paragraph{Fixed action set example.} For given $d$ and $\mu$, we sample $\Param\sim\Normal{\mu\One_{3d},\Eye_{3d}}$. Then, we reveal the action set $\ArmSet_t=\{0,\Arm,\Arm'\}$ as defined in \cref{eq:fixed-arm-set}. Then, conditional on $\ChosenArm_t=\Arm'$, we compute the probability $p$ that the next arm is not $0$. Also, recall from \cref{subsec:example-fixed-action-set} that, under complement of this event, when \LinTS{} incorrectly chooses action $0$, this probability does not change in the next round. Hence, \LinTS{} would be expected to fail for $1/p$ time periods.
We repeat this process 100 times to get $(p_i)_{i=1}^{100}$, and as before, we show a boxplot for each value of the varying variable. \cref{fig:failures}(b) shows the boxplots of $1/p_i$ for $\mu=0.1$ when $d$ varies between $1$ to $2^{18}$, and \cref{fig:failures}(c) illustrates $1/p_i$ for $d=1000$ and $\mu$ varying between 0 and 1. The results validate the theoretical analysis of \cref{sec:brittleness}.

\subsection{Thinness over time and \TSAI{}}
\label{sec:simulations-thinness}

In this subsection, we present two sets of simulations. Firstly, we investigate the variation of the thinness parameter over time in \cref{sec:improved-ts} and then compare the performance of \TSAI{} with both the Bayesian and frequentist versions of \LinTS{} in two different scenarios. The first scenario is a ``well-behaved" setting where the Bayesian \LinTS{} does not fail. The second scenario, known as Examples 1 and 2 in \cref{sec:brittleness}, are the brittle settings where Bayesian \LinTS{} fails.

\paragraph{Scenario I.} 
We consider a setting similar to the simulations section of \citep{russo2014learning}. Specifically, for $d=50$, we generate the parameter vector $\Param$ from the standard normal distribution $\Normal{0,10\Eye_d}$. At each time step $t$, we generate $k=100$ independent vectors from the uniform distribution on the hypercube $[-1/\sqrt{d},1/\sqrt{d}]^d$, which form the action set. We compare the following policies:
\begin{enumerate}
\item TS-Bayes: \cref{alg:proper-ts} with no inflation ($\Inflation=1$). 
\item TS-Freq: \cref{alg:proper-ts} with $\Inflation=\Radius_t$ at time $t$. This is the version considered by \cite{agrawal2013thompson} and \cite{abeille2017linear}.
\item \TSAI{}: \cref{alg:tsai} with $\Inflation=5$ and $\ThinnessBound = 2.0$.
\end{enumerate}
Note that for both TS-Freq and \TSAI{}, we use
\begin{align}
    \Radius_t
    :=
    \sqrt{2\log\left(\frac{\det(\SymCovMatrix_t^{-1})^{\frac12}\det(0.1\Eye_d)^{-\frac12}}{0.0001}\right)}+\sqrt{d}\,.
    \label{eq:Radius}
\end{align}
\begin{figure}[h]
	\begin{center}
		\begin{subfigure}{0.49\textwidth}
			\includegraphics[width=\textwidth]{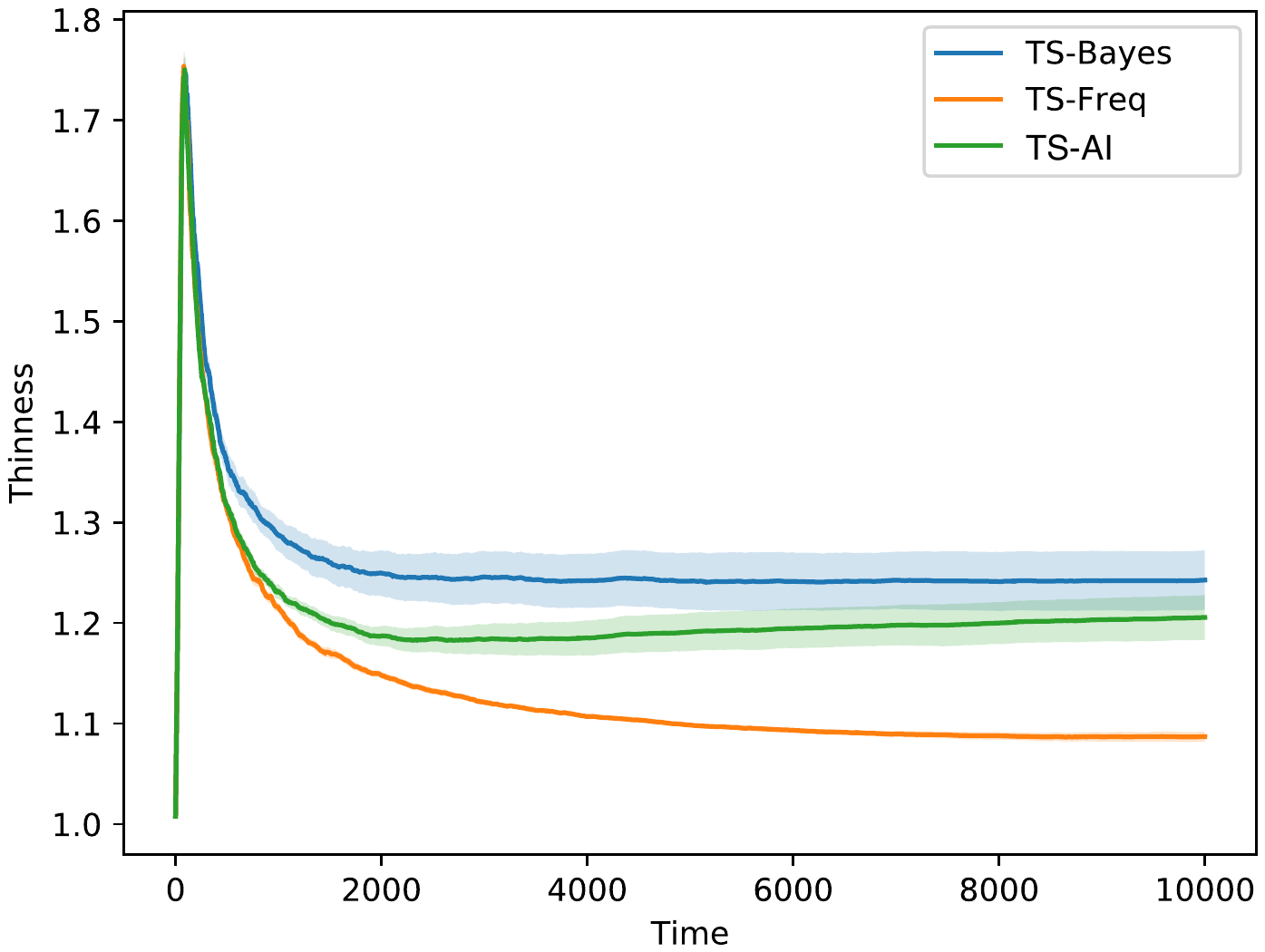}
			\caption{{\footnotesize Thinness over time}}
		\end{subfigure}
		\hfill
		\begin{subfigure}{0.49\textwidth}
			\includegraphics[width=\textwidth]{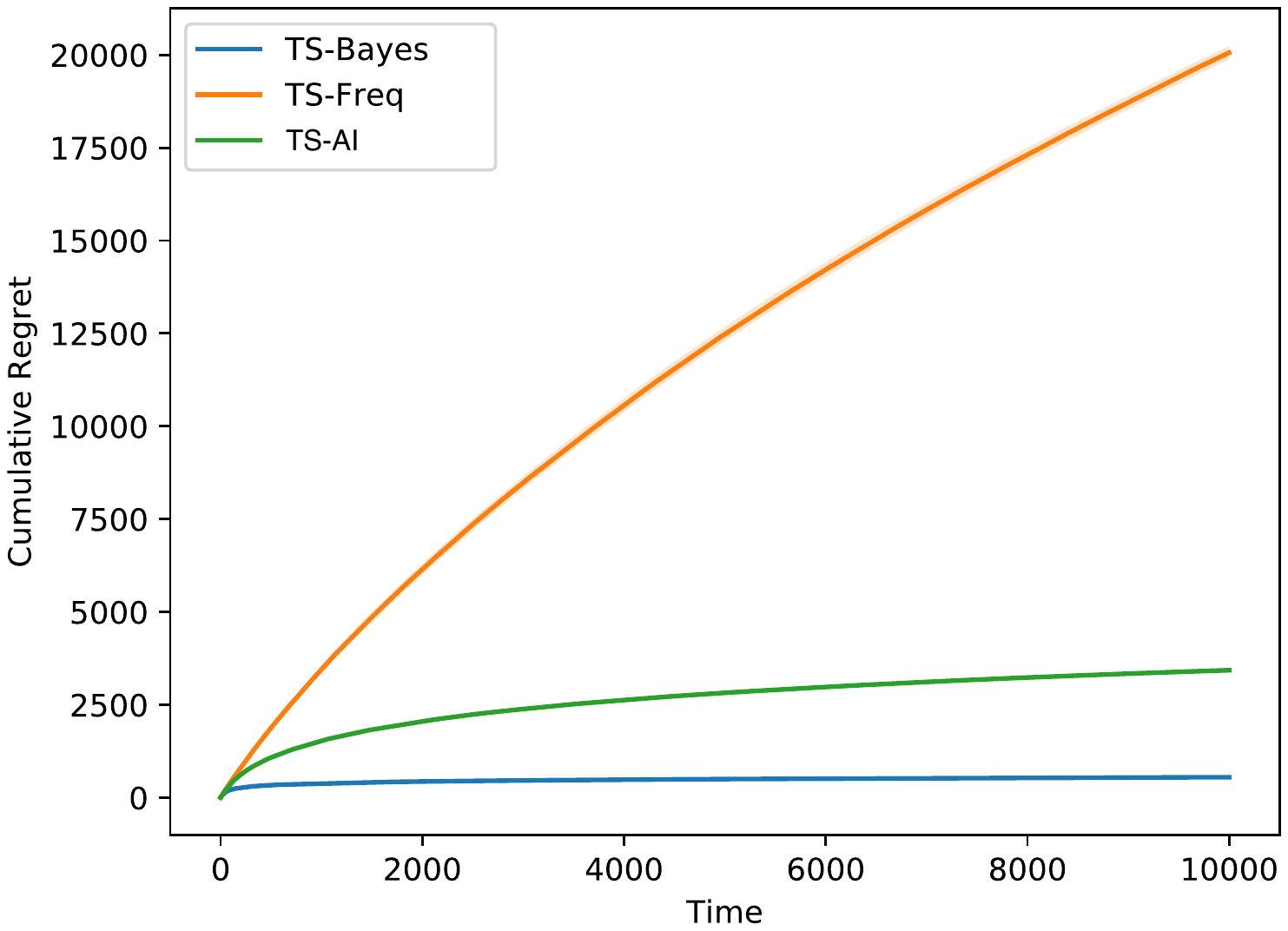}
			\caption{{\footnotesize Cumulative regret}}
		\end{subfigure}
		\vspace{3mm}
		\caption{Scenario I: Thinness and cumulative regret of \TSAI{} versus Bayesian and Frequentist versions of \LinTS{} in a well-behaved setting.}
		\label{fig:scnarioI}
	\end{center}
\end{figure}
%
Each policy chooses $\ChosenArm_t^\pi$ for $\pi\in\{\text{TS-Bayes, TS-Freq, TS-AI}\}$, and receives feedback $Y_t^\pi=\Inner[\big]{\Param,\ChosenArm_t^\pi}+\eps_t^\pi$ where $\eps_t^\pi$ are i.i.d. standard Gaussian random variables. Next, we compute the thinness parameter for $\SymCovMatrix_t^\pi=\Eye/10+\sum_{j=1}^{t}\ChosenArm_j^\pi\ChosenArm_j^{\pi\top}$. We repeat this procedure 20 times. \cref{fig:scnarioI}(a) displays the thinness of these policies in our experiments. This in particular shows that the thinness stays close to 1 for larger values of $t$. In other words, the term $\sum_{t=1}^{T}\Indicator(\Thinness(\SymCovMatrix_t)>\ThinnessBound)$ as it appears in \cref{thm:lin-ts-regret-informal} (and its formal version, \cref{thm:lin-ts-regret}) is zero for $\ThinnessBound<2$ with high probability. \cref{fig:scnarioI}(b) shows the cumulative regrets of these policies. Notice that, while \TSAI{} may inflate the posterior variance by $\Radius_t$, its performance is closer to TS-Bayes than TS-Freq given that the decision to inflate is performed in a more data-driven fashion.

\begin{figure}[h]
	\begin{center}
		\begin{subfigure}{0.49\textwidth}
			\includegraphics[width=\textwidth]{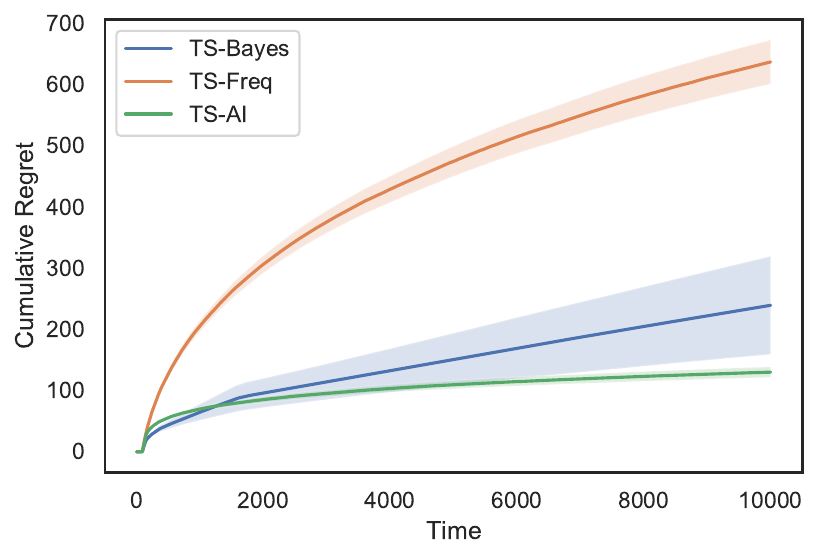}
			\caption{{\footnotesize Example 1}}
		\end{subfigure}
		\hfill
		\begin{subfigure}{0.49\textwidth}
			\includegraphics[width=\textwidth]{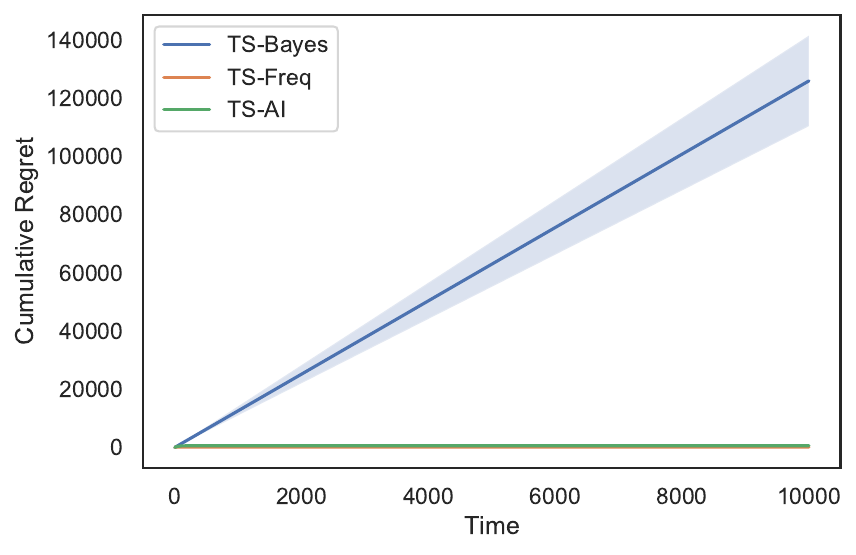}
			\caption{{\footnotesize Example 2}}
		\end{subfigure}
		\vspace{3mm}
		\caption{Scenario II: Thinness and cumulative regret of \TSAI{} versus Bayesian and Frequentist version of \LinTS{} in the setting of Examples 1 and 2 from \cref{sec:brittleness}.}
		\label{fig:scen2}
	\end{center}
\end{figure}

\paragraph{Scenario II.} Here, we consider the settings of Examples 1 and 2 from \cref{sec:brittleness}. For Example 1, we choose $d=90$ and the only difference is that we select $\sigma^2=1$ while $\tau^2=2$. For Example 2, we choose $d=30$, there is no variance mismatch ($\sigma^2=\tau^2=2$), but there is mean mismatch (prior mean is $10$ while \LinTS{} assumes prior mean is $0$). In each case,  we repeat the simulation 100 times and show the average regrets, with shaded error bars representing two standard errors, in \cref{fig:scen2}.  In both cases, TS-Bayes, as predicted performs poorly but \TSAI{} nearly ties or outperforms TS-Freq, benefiting from the adaptive inflation.

\section{Justifying \cref{as:optimal-diversity,as:optimal-random-friendly}  in a Concrete Example}
\label{sec:constants}

In this section, we prove that parameters $\DivParam$ and $\ExploreRatio$
in \cref{as:optimal-diversity,as:optimal-random-friendly} are constants, in the specific example from \cite{russo2014learning} that was empirically studied in \cref{sec:simulations-thinness}. 

Proofs of 
\cref{lem:single-random-friendly,lem:single-arm-diversity} are given in \cref{sec:auxi}.

Let $k$ be the number of actions at each round. We first start by verifying \cref{as:optimal-random-friendly}.
\begin{lem}
\label{lem:single-random-friendly}
Let $\Arm$ be sampled from $\Uniform([-1/\sqrt{d},1/\sqrt{d}]^d)$. Then, for any positive definite $\SymCovMatrix$ with $\Thinness(\SymCovMatrix)\leq\ThinnessBound$, we have that
\begin{align*}
    \Prob*{
        \Abs[\Big]{\Norm{\Arm}_{\SymCovMatrix}^2-\frac1{3\Dim}\NormNuc{\SymCovMatrix}}
        \leq
        \frac1{6\Dim}\NormNuc{\SymCovMatrix}
    }
    \leq
    2\exp\left(
        -\frac{c\Dim}{\ThinnessBound^2}
    \right)
\end{align*}
where $c$ is an absolute constant.
\end{lem}
The following corollary is a direct consequence of the above lemma combined with the union bound.
\begin{cor}
\label{cor:optimal-random-friendly}
Let $\OptimalArm_t$ be the optimal action at time $t$. Then, for any positive definite $\SymCovMatrix$ with $\Thinness(\SymCovMatrix)\leq\ThinnessBound$, we have that
\begin{align*}
    \Prob*{
        \Abs[\Big]{\Norm{\OptimalArm_t}_{\SymCovMatrix}^2-\frac1{3\Dim}\NormNuc{\SymCovMatrix}}
        \leq
        \frac1{6\Dim}\NormNuc{\SymCovMatrix}
    }
    \leq
    2k\cdot\exp\left(
        -\frac{c\Dim}{\ThinnessBound^2}
    \right).
\end{align*}
Specifically, by setting $\SymCovMatrix:=\Eye_\Dim$, we get that
\begin{align}
    \Prob*{
        \Norm{\OptimalArm_t}_2^2\leq\frac16
        ~~~~~\text{or}~~~~~
        \Norm{\OptimalArm_t}_2^2\geq\frac12
    }
    \leq
    2k\cdot\exp\left(
        -c\Dim
    \right),
    \label{eq:random-friendly-eye}
\end{align}
and hence
\begin{align}
    \Prob*{
        \Norm{\OptimalArm_t}_{\SymCovMatrix}^2
        \geq
        \frac1{3\Dim}\NormNuc{\SymCovMatrix}\cdot\Norm{\OptimalArm_t}_2^2
    }
    \leq
    4k\cdot\exp\left(
        -c\Dim
    \right),
    \label{eq:random-friendly-concrete}
\end{align}
\end{cor}
Our next lemma asserts that each action satisfies \cref{as:optimal-diversity} with a constant parameter $\DivParam$.
\begin{lem}
\label{lem:single-arm-diversity}
Let $\Arm$ be chosen according to $\Uniform([-1/\sqrt{d},1/\sqrt{d}]^d)$. Then, for any $V\in\IR^d$ and $p\in(0,1)$, we have
\begin{align*}
    \Prob*{\Inner{\Arm,V} > \sqrt{\frac{2\log(1/p)}{\Dim}}}
    \leq
    p.
\end{align*}
Furthermore, using \cref{eq:random-friendly-eye}, we get that
\begin{align*}
    \Prob*{
        \Inner{\Arm,V} > \sqrt{\frac{12\log(1/p)}{\Dim}}\cdot\Norm{\Arm}_2
    }
    \leq
    p+2k\cdot\exp\left(-c\Dim\right).
\end{align*}
\end{lem}
By applying the union bound, we can obtain the following result for the optimal arm. This corollary directly follows from the previous lemma and the application of the union bound.
\begin{cor}
\label{cor:optimal-arm-diversity}
Let $\OptimalArm_t$ be the optimal arm at time $t$. Then, for any $V\in\IR^d$, we have that
\begin{align*}
    \Prob*{
        \Inner{\OptimalArm_t,V} > \sqrt{\frac{12\log(2kT^2)}{\Dim}}\cdot\Norm{\OptimalArm_t}_2
    }
    \leq
    \frac{1}{2T^2}+2k\cdot\exp\left(-c\Dim\right)\,.
\end{align*}
\end{cor}
This means, if $\Dim$ is larger than
$\ThinnessBound^2\log(8kT^2)/c$,
\cref{as:optimal-diversity,as:optimal-random-friendly} are satisfied with
\begin{align*}
    \DivParam
    :=
    \sqrt{12\log(2kT^2)}
    ~~~~~\text{and}~~~~~
    \ExploreRatio
    :=
    \frac{1}{\sqrt{3}}.
\end{align*}
\section{Conclusion and Discussion}
\label{sec:discussion}

This paper focuses on the stochastic linear bandit problem and investigates the Linear Thompson Sampling (LinTS) algorithm. Our goal is to determine if the factor $d$ inflation in the posterior variance of LinTS, necessary to achieve the best-known frequentist regret bound, is essential. By settling an open problem, we show that the factor $d$ inflation is indeed necessary and that the frequentist regret bound of $\widetilde{\mathcal{O}}(d\sqrt{dT})$ is optimal. Additionally, we demonstrate that more data-driven versions of LinTS, which use the observed data to adjust the posterior inflation, can achieve frequentist minimax optimal regret under additional conditions.

While our main results are theoretical, our paper provides insights into the performance of LinTS and identifies potential sources of degradation that may be of interest for practitioners. Our analysis highlights that a even a small mismatch between the prior and true distributions can lead to suboptimal performance, which supports prior literature and emphasizes the importance of careful prior distribution selection. Furthermore, we find that the randomization bias arising from inherent ``sampling'' nature of the algorithm can be a potential source of degradation in the performance of LinTS. Further understanding the impact of this bias is an intriguing topic for future research, and we hope that our findings will encourage further exploration in this area.

\begin{APPENDICES}

\section{Proofs of \cref{sec:brittleness}}
\label{sec:brittleness-proofs}

Prior to commencing the proof, we shall present several fundamental definitions and properties concerning Gaussian and sub-Gaussian random variables. A more thorough discussion on this topic can be found in \citep{vershynin2018high}.

The sub-Gaussian norm of a random variable $X$ by
\[
\NormSubG{X} =
\inf\left\{t>0:~\Expect[\big]{e^{\left(\frac{X}{t}\right)^2}}\leq2\right\}\,.
\]
For every sub-Gaussian random variable $X$, there exist positive constants
$C_{\text{tail}}$ and $C_{\text{mgf}}\,$, such that
\begin{align}
	\Prob*{X\geq |t|}&\leq 
	2\,\exp{(-\frac{C_{\text{tail}}\,t^2}{\NormSubG{X}^2})}\,.
	\label{eq:SubGaussianTail}\\
	\Expect{\exp{(s(X-\Expect{X})}}&\leq \exp{(C_{\text{mgf}}\, s^2 \NormSubG{X}^2)}\,,~~~~\text{for all $s\in\IR$}\,,
\label{eq:SubGaussianMGF}	
\end{align}
and the Gaussian distribution satisfies
\begin{align}
	\NormSubG{\Normal{0,\sigma^2}}&\leq 2\sigma\,,
	\label{eq:GaussianNormSubG}\\
	\Prob[\Big]{\Normal{0,\sigma^2}\geq |t|}&\leq 
\frac{1}{\sqrt{2\pi}}\frac{\sigma}{|t|}\exp{(-\frac{t^2}{2\sigma^2})}\,.
\label{eq:GaussianTailSharp}	
\end{align}
We also need the following proposition that is proved in  \cref{sec:auxi}. 
\begin{prop}[Bias decomposition]
	Let $(X_i)_{i=1}^n$ be a sequence of independent random variables where $X_i\sim\Normal{0,\sigma_i^2}$. By $Y$, we denote their sum and let $Z$ be any independent random variable. Then, for any function $g:\IR\times\IR\to\IR$, we have
	\begin{align*}
		\Expect*{X_i\cdot g(Y,Z)}
		=
		\frac{\sigma_i^2}{\sum_{i=1}^{n}\sigma_i^2}\cdot\Expect*{Y\cdot g(Y,Z)}.
	\end{align*}
	\label{prop:bias-decomp}
\end{prop}

\begin{proof}[Proof of \cref{lem:bias-quant}]	
Recall that, as stated in \cref{rem:example-1-guide}, for notation simplicity, $\lambda=1$. It follows from the definition of $\Estimator_3$ that
\begin{align*}
    \Estimator_3
    =\frac12\begin{bmatrix}
        \RewardValue_1\\
        \RewardValue_2
    \end{bmatrix}
    =\frac12\begin{bmatrix}
        \Param_1+\eps_1\\
        \Param_2+\eps_2
    \end{bmatrix}.
\end{align*} 
Next, at $t=3$, the $i$-th entry is updated according to
\begin{align*}
    \Estimator_{4,i}
    &=
    \frac{\RewardValue_i+\RewardValue_3}3\\
    &=
    \frac{2\Param_i+\eps_i+\eps_3}3\\
    &=
    \frac{\Param_i+\eps_i}2+\frac{\Param_i-\eps_i}{6}+\frac{\eps_3}3\\
    &=
    \Estimator_{3,i}+\frac{\Param_i-\eps_i}{6}+\frac{\eps_3}3.
\end{align*}
Moreover, the other entry remains unchanged. In other words,
\begin{align*}
    \Estimator_{4,3-i}=\Estimator_{3,3-i}\,.
\end{align*}
Therefore, setting $V=e_1+e_2$, we have
\begin{align*}
    \Inner[\big]{\Estimator_4,V}
    &=
    \Inner[\big]{\Estimator_4,V}\\
    &=
    \Estimator_{4,1}+\Estimator_{4,2}\\
    &=
    \Estimator_{3,1}+\Estimator_{3,2}+\frac{\Param_i-\eps_i}{6}+\frac{\eps_3}3.\Yesnumber
    \label{eq:v-prod}
\end{align*}
and in particular
\begin{align}
    \Inner[\big]{\Expect[\big]{\Estimator_4},V}
    &=
    \Expect*{\frac{\Param_i-\eps_i}{6}}.
    \label{eq:v-bias}
\end{align}
We can now compute this expression in terms of the \emph{\randombias{} bias coefficient} given by
\begin{align*}
    \BiasCoeff:=\Expect*{\max\{A, B\}}>0,
\end{align*}
where $A$ and $B$ are two independent standard normal random variables. Our main tool in this calculation is the bias decomposition, stated in \cref{prop:bias-decomp}. Recall that
\begin{align*}
    i=\Argmax_{j\in[1,2]}\,\TsSample_{3,j}.
\end{align*}
By definition,
\begin{align*}
\TsSample_{3}\sim\Normal[\Big]{0,\left(\frac{\sigma^2+\tau^2+2}{4}\right)\II_2}\,.
\end{align*}
Therefore, we have
\begin{align*}
    \Expect*{\TsSample_{3,i}}
    &=
    \sqrt{\frac{\sigma^2+\tau^2+2}{4}}\cdot\BiasCoeff.
\end{align*}
On the other hand, it follows from the symmetry that
\begin{align}
    \Expect*{\TsSample_{3,i}}
    &=
    2\Expect*{\TsSample_{3,1}\cdot\II(i=1)}\nonumber\\
    &=
    2\Expect*{\TsSample_{3,1}\cdot\II(\TsSample_{3,1}\geq\TsSample_{3,2})}\,.
    	\label{eq:BiasTsSample}
\end{align}
Combining \cref{prop:bias-decomp} for the sequence
\begin{align*}
    X_1:=\frac{\Param_1}2,
    ~~~~~~~~
    X_2:=\frac{\eps_1}2,
    ~~~~~~~~
    \text{and}
    ~~~~~~~~
    X_3:=\TsSample_{3,1}-\Estimator_{3,1 }\sim\Normal*{0,\frac12}\,,
\end{align*}
and $Z=\TsSample_{3,2}$, with \cref{eq:BiasTsSample}, we infer that
\begin{align*}
    \Expect*{\frac{\Param_1}2\cdot\II(\TsSample_{3,1}\geq\TsSample_{3,2})}
    &=
    \frac{\sigma^2}{\sigma^2+\tau^2+2}\Expect*{\frac{\TsSample_{3,1}}2\cdot\II(\TsSample_{3,1}\geq\TsSample_{3,2})}\\
    &=
\frac{\sigma^2}{\sigma^2+\tau^2+2}\cdot\frac{\sqrt{\sigma^2+\tau^2+2}}{4}\cdot\BiasCoeff\\    
    &=
    \frac{\sigma^2\BiasCoeff}{4\sqrt{\sigma^2+\tau^2+2}}\,.
\end{align*}
Consequently, we can write
\begin{align*}
    \Expect*{\Param_i}
    &=
    2\,\Expect*{\Param_1\cdot\II(\TsSample_{3,1}\geq\TsSample_{3,2})}\\
    &=
    4\,\Expect*{\frac{\Param_1}2\cdot\II(\TsSample_{3,1}\geq\TsSample_{3,2})}\\
    &=
    \frac{\sigma^2\BiasCoeff}{\sqrt{\sigma^2+\tau^2+2}}.\Yesnumber
    \label{eq:theta-i-bias}
\end{align*}
Similarly, we can conclude that
\begin{align}
    \Expect*{\eps_i}
    &=
    \frac{\tau^2\BiasCoeff}{\sqrt{\sigma^2+\tau^2+2}}.
    \label{eq:eps-i-bias}
\end{align}
Combining \cref{eq:v-bias} with \cref{eq:theta-i-bias} and \cref{eq:eps-i-bias}, we obtain
\begin{align*}
    \Inner[\big]{\Expect[\big]{\Estimator_4},V}
    &=
    \frac{\left(\sigma^2-\tau^2\right)\BiasCoeff}{6\sqrt{\sigma^2+\tau^2+2}}.
\end{align*}
This equality implies that $\Estimator_4$ is directionally biased whenever $\sigma^2\neq\tau^2$. Finally, \cref{eq:GaussianNormSubG} and \cref{eq:v-prod} give
\begin{align*}
    \NormSubG[\big]{\Inner[\big]{\Estimator_4,V}}
    &=
    \NormSubG[\big]{\Estimator_{3,1}+\Estimator_{3,2}+\frac{\Param_i-\eps_i}{6}+\frac{\eps_3}3}\\
    &\leq
    \NormSubG[\big]{\Estimator_{3,1}}+\NormSubG[\big]{\Estimator_{3,2}}+\frac16\NormSubG[\big]{\Param_i}+\frac16\NormSubG[\big]{\eps_i}+\frac13\NormSubG[\big]{\eps_3}\\
    &\leq
    2\sqrt{\sigma^2+\tau^2+2}+\frac16\NormSubG[\big]{\Param_i}+\frac16\NormSubG[\big]{\eps_i}+\frac{2\tau}{3}\,.
\end{align*}
Noting that, and using \cref{eq:GaussianNormSubG} again,
\begin{align*}
    \NormSubG[\big]{\Param_i}
    =
    \NormSubG[\big]{\Abs*{\Param_i}}
    \leq
    \NormSubG[\big]{\Abs*{\Param_1}+\Abs*{\Param_2}}
    \leq
    2\NormSubG[\big]{\Abs*{\Param_1}}
    \leq 4\sigma
\end{align*}
and similarly for $\eps_i$ we can show $ \NormSubG[\big]{\eps_i}\le 4\tau$ which means that
\begin{align*}
    \NormSubG[\big]{\Inner[\big]{\Estimator_4,V}}
    &\leq
   2 \sqrt{\sigma^2+\tau^2+2}+\frac23(\sigma+\tau)+\frac{2\tau}{3}\\
    &\leq
    4(\sigma+\tau+\sqrt{2}).
\end{align*}
Therefore, we have
\begin{align*}
    \NormSubG[\big]{\Inner[\big]{\Estimator_4,V}-\Expect[\big]{\Inner[\big]{\Estimator_4,V}}}
    &\leq
    \NormSubG[\big]{\Inner[\big]{\Estimator_4,V}}+\NormSubG[\big]{\Expect[\big]{\Inner[\big]{\Estimator_4,V}}}\\
    &\leq
    8(\sigma+\tau+\sqrt{2})\,.
\end{align*}
This and \ref{eq:SubGaussianMGF} imply that the m.g.f. of 
$\Estimator_4-\Expect[\big]{\Estimator_4}$ satisfies
\begin{align*}
    \Expect*{\exp\left(s\Inner[\big]{V,\Estimator_4-\Expect[\big]{\Estimator_4}}\right)}
    \leq
    \exp\left[64C_{\text{mgf}}\,s^2 (\sigma+\tau+\sqrt{2})^2\right]\,,
    ~~~~
    \text{for all $s\in\IR$.}
\end{align*}
\end{proof}
\begin{proof}[Proof of \cref{lem:misperception}]
Since all the $d$ blocks are decoupled, it follows from \cref{eq:v-bias-closed-form} that
\begin{align}
    \Expect*{\Inner[\big]{\Estimator_{3d+1},\Arm}}=-C_1(\sigma,\tau)\sqrt d
    \label{eq:bias-sum}
\end{align}
where \[C_1(\sigma,\tau)=\frac{\Abs*{\sigma^2-\tau^2}\cdot\BiasCoeff}{6\sqrt{\sigma^2+\tau^2+2}}\,.
\]
Assuming $\sigma^2\neq\tau^2$, we observe that $C_1(\sigma,\tau)>0$. Moreover, \cref{eq:bias-mgf} implies that
\begin{align*}
    \Expect*{\exp\left(s\left(\Inner[\big]{\Estimator_{3d+1},\Arm}+C_1(\sigma,\tau)\sqrt d\right)\right)}
    \leq
    \exp\left[Cs^2(\sigma+\tau+\sqrt{2})^2\right],
    ~~~~
    \text{for all $s\in\IR$.}
\end{align*}
which means, for a constant $C'$,
\begin{align*}
    \NormSubG[\Big]{\Inner[\big]{\Estimator_{3d+1},\Arm}+C_1(\sigma,\tau)\sqrt d}
    &\leq
    C'(\sigma+\tau+\sqrt{2})\,.
\end{align*}
Using this inequality in combination with \cref{eq:SubGaussianTail},  and \cref{eq:bias-sum}, we assert the following concentration inequality
\begin{align*}
    \Prob*{\Inner[\big]{\Estimator_{3d+1},\Arm}\leq-\frac{C_1(\sigma,\tau)\sqrt d}{2}}
    &=
    \Prob*{\Inner[\big]{\Estimator_{3d+1},\Arm}+C_1(\sigma,\tau)\sqrt d\leq\frac{C_1(\sigma,\tau)\sqrt d}{2}}\\
    &\geq    1-2\exp\left[-C_{\text{tail}}\frac{C_1(\sigma,\tau)^2d}{4C'^2(\sigma+\tau+\sqrt{2})^2}\right]\,.  
\end{align*}
Next, note that $\Inner*{\Param,\Arm}\sim\Normal*{0,2\sigma^2}$, and thus, we have
\begin{align*}
    \Prob*{\Inner[\big]{\Param,\Arm}\geq\sqrt 2\sigma}=1-\Phi(1)\,,
\end{align*}
and for sufficiently large values of $d$,
\[
2\exp\left[-C_{\text{tail}}\frac{C_1(\sigma,\tau)^2d}{4C'^2(\sigma+\tau+\sqrt{2})^2}\right]\leq \frac12\big(1-\Phi(1)\big)\,,
\]
and hence
\begin{align*}
    \Prob*{\Inner[\big]{\Param,\Arm}\geq\sqrt 2\sigma~~\text{and}~~\Inner[\big]{\Estimator_{3d+1},\Arm}\leq-\frac{C_1(\sigma,\tau)\sqrt d}{2}}\geq p_0\,.
\end{align*}
\end{proof}

\begin{proof}[Proof of \cref{prop:noise-reduction-failure}]
First, note that the regret for each block is of order $1$ because each of the two actions is equally likely to be selected.  Therefore, the regret during the first $3d$ periods is of order $d$. This means, unless  $d=o(T)$, the regret would already be linear in $T$. Therefore, in the remaining we assume $d=o(T)$.

For $t>3d$, let $Z_t$ be given by
\begin{align*}
    Z_t:=\begin{cases}
        1 & \text{if action $\Arm$ is never selected up to time $t$,}\\
        0 & \text{otherwise.}
    \end{cases}
\end{align*}
We now have the following lower bound for the regret of \cref{alg:proper-ts}:
\begin{align*}
    \Regret(T,\PolicyTS,\Prior)
    &\geq
    \Prob*{\BadEvent}\cdot\Expect*{\sqrt 2\sigma\cdot\sum_{t=3d+1}^{T}Z_t\Given\BadEvent}\\
    &\geq
    \sqrt 2\sigma p_0\cdot\sum_{t=3d+1}^{T}\Expect*{Z_t\Given\BadEvent}\\
    &=
    \sqrt 2\sigma p_0\cdot\sum_{t=3d+1}^{T}\Prob*{Z_t\Given\BadEvent}.
\end{align*}
Define $q:=\Prob*{1-Z_{3d+1}\Given\BadEvent}$. We get that
\begin{align*}
    \Prob*{Z_t\Given\BadEvent}
    &=
    \Prob*{Z_{t}\Given\BadEvent,Z_{t-1}}\cdot\Prob*{Z_{t-1}\Given\BadEvent}\\
    &=
    (1-q)\cdot\Prob*{Z_{t-1}\Given\BadEvent}\\
    &=
    (1-q)^{t-3d}\,.
\end{align*}
because $Z_{t} \,|\, \BadEvent,Z_{t-1}$ has the same distribution as $Z_{3d+1}\,|\,\BadEvent$ since choosing action $0$ will not change \LinTS's posterior estimate.

Furthermore, it follows from the definition of $q$ and \cref{eq:GaussianTailSharp} that
\begin{align*}
    1-q
    &\geq
    \Prob*{\Inner[\big]{\TsSample_{3d+1}-\Estimator_{3d+1},\Arm}\leq\frac{C_1(\sigma,\tau)\sqrt d}{2}}\\
    &\geq
    \Prob*{\Normal*{0,1}\leq\frac{C_1(\sigma,\tau)\sqrt d}{2}}\\
    &\geq
    1-\exp\Big(-C_2(\sigma,\tau)d\Big)\,,
\end{align*}
for a positive constant $C_2(\sigma,\tau)$. By combining the above, we have that
\begin{align*}
    \Regret(T,\PolicyTS,\Prior)
    &\geq
    \sqrt 2\sigma p_0 \sum_{t=3d+1}^{T}\left[1-\exp\Big(-C_2(\sigma,\tau)d\Big)\right]^{t-3d}\\
    &=
    \sqrt 2\sigma p_0 \frac{1-\left[1-\exp\Big(-C_2(\sigma,\tau)d\Big)\right]^{t-3d}}{\exp\Big(-C_2(\sigma,\tau)d\Big)}\\
   &\geq  \sqrt 2\sigma p_0 (T-3d)\left(1-\frac{T-3d}{2}\exp\Big(-C_2(\sigma,\tau)d\Big)\right)\,, 
\end{align*}
where the last step uses inequality $1-(1-\alpha)^n\ge \alpha[1-(n-1)\alpha/2]$ for any  integer $n>2$ and $\alpha\in (0,1)$. This immediately follows that, whenever $T\leq \exp\Big(C_2(\sigma,\tau)d\Big)$,
\begin{align*}
    \Regret(T,\PolicyTS,\Prior)
    \geq
    \frac{\sqrt 2\sigma p_0}{2}(T-3d)\geq \LowerOrder{T}\,.
\end{align*}
In other words, the regret of \LinTS{} grows linearly up to time $\exp\Big(C_2(\sigma,\tau)d\Big)$.
\end{proof}

\begin{proof}[Proof of \cref{prop:mean-shift-failure}]
Notice that
\begin{align*}
    \Inner*{\Param,\Arm'}\sim\Normal*{-\mu\sqrt d,\sigma^2}
    ~~~~~\text{and}~~~~~
    \Inner*{\Param,\Arm}\sim\Normal*{\mu\sqrt d,3\sigma^2}.
\end{align*}
Therefore, for a positive constant $C$, $\Inner*{\Param,\Arm}\geq\frac12\mu\sqrt d$ and $\Inner*{\Param,\Arm'}\leq-\frac12\mu\sqrt d$ simultaneously with probability at least $1-2\exp\left(-C\frac{\mu^2d}{\sigma^2}\right)$. This thus implies that $\Arm$ is the optimal arm with high probability. For sufficiently large $d$, this probability exceeds $\frac78$.

On the other hand, at $t=1$, \LinTS{} (\cref{alg:proper-ts}) will choose $\Arm'$ with probability $\frac14$. This holds true as $\Arm'$ is chosen if and only if
\begin{align*}
    \Inner[\big]{\TsSample_1,\Arm'} > 0
    ~~~~~\text{and}~~~~~
    \Inner[\big]{\TsSample_1,\Arm-\Arm'} < 0.
\end{align*}
The claim follows from the fact these two random variables are two centered and independent normal random variables. In this case, we have
\begin{align*}
    \SymCovMatrix_2
    &=(\II_{3d}+\Arm'\Arm'^\top)^{-1}\\
    &= \II_{3d}-\frac{1}{2}\Arm'\Arm'^\top
    ~~~~~\text{and}~~~~~\\
    \Estimator_2 &=  \frac12\Arm'Y_1\\
   &= \frac12\Arm'(\Inner*{\Param,\Arm'}+\eps_1)\,.
\end{align*}
Next, we provide an upper bound for the probability that \LinTS{} chooses arm $0$ at $t=2$. This happens if and only if
\begin{align*}
    \Inner[\big]{\TsSample_2,\Arm'} < 0
    ~~~~~\text{and}~~~~~
    \Inner[\big]{\TsSample_2,\Arm} < 0.
\end{align*}
Note that
\begin{align}
    \Inner[\big]{\TsSample_2,\Arm'}\sim\Normal*{\frac{Y_1}2,\frac12}
    ~~~~~\text{and}~~~~~
    \Inner[\big]{\TsSample_2,\Arm}\sim\Normal*{\frac{Y_1}2,\frac52}.
    \label{eq:mean-shift-round-2-prob}
\end{align}
For sufficiently large $d$, we have
\begin{align*}
    \Prob*{\eps_1 > \frac14\mu\sqrt d}\leq\frac{1}{16}.
\end{align*}
From $Y_1=\Inner*{\Param,\Arm'}+\eps_1$ and the union bound it follows that $\Prob*{\BadEvent'}\geq 1/16$ where $\BadEvent'$ is defined by
\begin{align*}
\BadEvent':=\left\{\Inner*{\Param,\Arm}\geq\frac12\mu\sqrt d, ~~~~~~\Inner*{\Param,\Arm'}\leq-\frac12\mu\sqrt d, ~~~~~ \ChosenArm_1=\Arm',~~~\text{and}~~~ Y_1<-\frac14\mu\sqrt d\right\}\,.
\end{align*}
From \cref{eq:mean-shift-round-2-prob}, we can deduce that, if $ q
:=\Prob*{\ChosenArm_2\neq0\Given\BadEvent'}$,
\begin{align*}
    1-q
    &:=
    \Prob*{\ChosenArm_2=0\Given\BadEvent'}\\
    &~=
    \Prob*{\Inner[\big]{\TsSample_2,\Arm'} < 0
        ~~~\text{and}~~~
        \Inner[\big]{\TsSample_2,\Arm} < 0\Given\BadEvent'}\\
    &~\geq
    1-2\exp\left(-C\mu^2d\right)\,,
\end{align*}
for a positive constant $d$ for a positive constant $C$.
Applying the same argument as in proof of \cref{prop:noise-reduction-failure}, we get that
\begin{align}
    \Regret(T,\PolicyTS,\Prior)
    \geq
    \frac{\mu\sqrt d}{32}T\left[1-\frac{T}{2}\exp(-C_3d)\right]
    \label{eq:mean-shift-lower-bound}
\end{align}
where $C_3$ is a constant that only depends on $\mu$, $\sigma$, and $\tau$ (but not on $d$). Therefore, for $T\leq \exp(-C_3d)$, the regret is linear in $T$.

\end{proof}
\section{A Formal Regret Bound for \LinTS and Proofs of \cref{sec:improved-ts}}
\label{sec:improved-ts-proofs}
This section is dedicated to our formal analysis of \LinTS{} under our proposed conditions. This also allows us to show that the confidence set in OFUL can also be shrunk significantly under similar conditions. To do so, we adopt the framework in \cite{hamidi2020general} to state our results, however we make small changes compared to them. We start by explicitly stating the conditions that we introduced in \cref{sec:improved-ts}. We say that the problem is in a \emph{well-posed} condition at time $t$ if
\begin{align}
\begin{split}
    \Thinness(\SymCovMatrix_t)&\leq\ThinnessBound,\\
    \Norm{\OptimalArm_t}_{\SymCovMatrix_t}
    &\geq
    \ExploreRatio\sqrt{\frac{\NormNuc{\SymCovMatrix_t}}\Dim}\cdot\Norm{\OptimalArm_t}_2,\\
    \Abs[\big]{\Inner[\big]{\OptimalArm_t,\Estimator_t-\Param}}
    &\leq
    \frac{\DivParam}{\sqrt\Dim}
    \Norm{\OptimalArm_t}_2\cdot\Norm[\big]{\Estimator_t-\Param}_2,
    \label{eq:well-posed-def}
\end{split}
\end{align}
and by $\IWellPosed_t$ we denote the indicator function of the above event. Now, by a \emph{worth} function we mean a (stochastic) function $\Worth_t(\cdot)$ that, given the history $\HistoryPlus_{t}$, assigns a real number to each action in the action set $\ArmSet_t$ with the additional condition that whenever $\IWellPosed_t=1$
\begin{align}
    \Prob*{
        \sup_{\Arm\in\ArmSet_t}\frac{\Abs{\Worth_t(\Arm)-\Inner{\Arm,\Estimator_t}}}{\Norm{\Arm}_{\SymCovMatrix_t}}
        \leq
        \RadiusWorth
        \Given
        \History_t,\ArmSet_t
    }
    \geq
    1-\frac1{T^2}\label{eq:worth-function-def}
\end{align}
for some fixed $\RadiusWorth>1$. We also say that $\Estimator_t$ is in a \emph{typical} condition with respect to $\Param$ if
\begin{align}
	\Prob*{
		\sup_{\Arm\in\ArmSet_t}\frac{\Abs{\Inner{\Arm,\Param}-\Inner{\Arm,\Estimator_t}}}{\Norm{\Arm}_{\SymCovMatrix_t}}
		\leq
		\RadiusParam
		\Given
		\History_t,\ArmSet_t
	}
	\geq
	1-\frac1{T^2}
	\label{eq:typical-def}
\end{align}
for some $\RadiusParam>1$ whenever $\IWellPosed_t=1$. We also let $\ITypicalM_t$ be the indicator function for this event. Intuitively, $\RadiusParam$ determines the width of the confidence interval around each $\Inner{\Estimator_t,\Arm}$ so that it contains $\Inner{\Param,\Arm}$ for all $\Arm\in\ArmSet_t$ simultaneously with high probability. Next, we say that the worth function $\Worth_t(\cdot)$ is \emph{optimistic} for given realizations $\History_t$ and $\ArmSet_t$ if
\begin{align*}
    \Prob*{
        \sup_{\Arm\in\ArmSet_t}\Worth_t(\Arm)
        \geq
        \sup_{\Arm\in\ArmSet_t}\Inner{\Arm,\Param}
        \Given
        \History_t,\ArmSet_t
    }
    \geq
    \OptProb
\end{align*}
for $\OptProb>0$ whenever $\IWellPosed_t=1$ and $\ITypicalM_t=1$. This notion of optimism \emph{cannot} hold almost surely as, for instance, $\Param-\Estimator_t$ can be arbitrarily large.

Using these notations, we introduce a modified version of Randomized OFUL (ROFUL), introduced in \cite{hamidi2020general}, that we analyze in this paper. The pseudo-code for this meta algorithm is presented in \cref{alg:roful}. Whenever $\Thinness(\SymCovMatrix_t)>\ThinnessBound$, ROFUL is allowed to select any arbitrary action. A natural choice is to choose the action that decreases $\Thinness(\SymCovMatrix_t)$ the most. An alternative is to define $\ChosenArm_t\gets \Argmax_{\Arm\in\ArmSet_t}\Worth_t(\Arm)$ as in Line 4. In this case, if one sets $\Worth_t(\Arm)=\Inner{\Arm,\TsSample_t}$ or $\Worth_t(\Arm)=\Inner{\Arm,\Estimator_t}+\RadiusWorth\Norm{\Arm}_{\SymCovMatrix_t}$, ROFUL becomes \LinTS{} or OFUL respectively. Because, for OFUL,  such $\Worth_t(\Arm)$ satisfiy \cref{eq:worth-function-def}, by definition. For  \LinTS{}, noting that $\Inner{\Arm,\TsSample_t}=\Inner{\Arm,\Estimator_t}+\Inflation\Norm{\Arm}_{\SymCovMatrix_t}\Normal{0,1}$, from \ref{eq:GaussianTailSharp} follows that \cref{eq:worth-function-def} holds as long as $\RadiusWorth/\Inflation = \LowerOrder{\sqrt{\log T}}$.

\begin{algorithm}[t]
	\caption{Randomized OFUL (ROFUL)}
	\label{alg:roful}
	\begin{algorithmic}[1]
		\REQUIRE Worth functions $(\Worth_t)_{t=1}^T$
		\FOR{$t=1,2,\cdots$}
		\STATE Observe $\ArmSet_t$
		\IF{$\Thinness(\SymCovMatrix_t)\leq\ThinnessBound$}
		\STATE $\ChosenArm_t\gets \Argmax_{\Arm\in\ArmSet_t}\Worth_t(\Arm)$
        \ELSE
        \STATE Choose any $\ChosenArm_t\in\ArmSet_t$
        \ENDIF
		\STATE Observe reward $Y_t$
		\ENDFOR
	\end{algorithmic}
\end{algorithm}

The next two lemmas assert that the optimism holds for \LinTS{} and OFUL. But we first recall the definition of $\Radius$ from \cite{abbasi2011improved}. 
\begin{align}
	\Radius:=3\sigma\sqrt{d\log\left(1+\lambda T \ArmBound^2\right)} + \lambda^{-1/2}\ParamBound\,,
	\label{eq:radius-def}
\end{align}
where $\ParamBound$ is an upper bound for $\Norm{\Param}_2$ and $\ArmBound$ is defined in \cref{sec:setting}. Therefore, $\rho = \OrderLog {\sqrt{d}}$. Also, 
Theorem 1 of \cite{abbasi2011improved} gives 
\begin{align}
	\Prob*{
		\Norm{\Param-\Estimator_{t}}_{\SymCovMatrix_t^{-1}}
		\leq
		\Radius
	}
	\geq
	1-\frac{1}{T^3}\,. \label{eq:oful-confidence-inequality}
\end{align}
Let $\ITypical_t$ be the binary indicator for the event in \cref{eq:oful-confidence-inequality}. Note that by Cauchy-Schwartz inequality one can easily see that if $\RadiusParam\ge\Radius$, then $\ITypicalM_t\ge \ITypical_t$. In fact, in the rest of this section, one can harmlessly assume $\RadiusParam=\Radius$ and only work with $\ITypical_t$. We only use separate notation to allow the possibility of $\RadiusParam<\Radius$.
\begin{lem}[Optimism of \LinTS{}]
\label{lem:lin-ts-optimism}
Set the inflation parameter of \LinTS{} to be 
\[
\Inflation:=\frac{\DivParam\ThinnessBound}{\ExploreRatio}\cdot\frac{\Radius}{\sqrt{d}} 
\]
and let $\Worth_t(\Arm):=\Inner{\Arm,\TsSample_t}$. Whenever $\IWellPosed_t=1$,  $\ITypical_t=1$, and $\ITypicalM_t=1$, we have
\begin{align}
    \Prob*{
        \sup_{\Arm\in\ArmSet_t}\Worth_t(\Arm)
        \geq
        \sup_{\Arm\in\ArmSet_t}\Inner[\big]{\Arm,\Param}
        \Given
        \History_{t},\ArmSet_t
    }
    \geq
    \Phi(-1).
    \label{eq:lin-ts-optimism}
\end{align}
\label{lem:lins-ts-optimism}
\end{lem}
\begin{lem}[Optimism of OFUL]
\label{lem:oful-optimism}
Set
$\Inflation$ as in \cref{lem:lins-ts-optimism} and $\Worth_t(\Arm):=\Inner{\Arm,\Estimator_t}+\Inflation\Norm{\Arm}_{\SymCovMatrix_t}$. Whenever $\IWellPosed_t=1$,  $\ITypical_t=1$, and $\ITypicalM_t=1$, we have
\begin{align}
    \Prob*{
        \sup_{\Arm\in\ArmSet_t}\Worth_t(\Arm)
        \geq
        \sup_{\Arm\in\ArmSet_t}\Inner[\big]{\Arm,\Param}
        \Given
        \History_{t},\ArmSet_t
    }
    =
    1.
    \label{eq:oful-optimism}
\end{align}
\end{lem}
We establish the proof of \cref{lem:lin-ts-optimism}, noting that the proof of \cref{lem:oful-optimism} would be almost identical and marginally simpler.
\begin{proof}[Proof of \cref{lem:lin-ts-optimism}]
We have
\begin{align*}
    \Inner[\big]{\TsSample_t,\OptimalArm_t}
    -
    \Inner[\big]{\Param,\OptimalArm_t}
    &=
    \Inner[\big]{\TsSample_t-\Estimator_t,\OptimalArm_t}
    -
    \Inner[\big]{\Param-\Estimator_t,\OptimalArm_t}\\
    &\geq
    \Inner[\big]{\TsSample_t-\Estimator_t,\OptimalArm_t}
    -
    \frac{\DivParam}{\sqrt d}
    \Norm{\OptimalArm_t}_2\cdot\Norm[\big]{\Estimator_t-\Param}_2\\
    &\geq
    \Inner[\big]{\TsSample_t-\Estimator_t,\OptimalArm_t}
    -
    \DivParam\Radius\sqrt{\frac{\NormOp{\SymCovMatrix_t}}{d}}\cdot\Norm{\OptimalArm}_2\,,
\end{align*}
where we used the third equation in \cref{eq:well-posed-def} first and then \cref{eq:oful-confidence-inequality}, combined with $\Norm{V}_2\leq \Norm{V}_{\SymCovMatrix_t^{-1}} \sqrt{\NormOp{\SymCovMatrix_t}}$ for any vector $V$.

Now, since  $\Inner[\big]{\TsSample_t-\Estimator_t,\OptimalArm_t}\sim\Normal{0,\Inflation^2\Norm{\OptimalArm}_{\SymCovMatrix_t}^2}$, we have
\begin{align*}
    \Prob*{
        \Inner[\big]{\TsSample_t,\OptimalArm_t}
        \geq
        \Inner[\big]{\Param,\OptimalArm_t}
    }
    &\geq
    \Prob*{
        \Inner[\big]{\TsSample_t-\Estimator_t,\OptimalArm_t}
        \geq
        \DivParam\Radius\sqrt{\frac{\NormOp{\SymCovMatrix_t}}{d}}\cdot\Norm{\OptimalArm}_2
    }\\
    &=
    \Phi\left(-
        \frac
        {\DivParam\Radius\sqrt{\frac{\NormOp{\SymCovMatrix_t}}{d}}\cdot\Norm{\OptimalArm}_2}
        {\Inflation\Norm{\OptimalArm}_{\SymCovMatrix_t}}
    \right)\\
    &\geq
    \Phi\left(-
        \frac
        {\DivParam\Radius\sqrt{\frac{\NormOp{\SymCovMatrix_t}}{d}}\cdot\Norm{\OptimalArm}_2}
        {\Inflation\ExploreRatio\sqrt{\frac{\NormNuc{\SymCovMatrix_t}}d}\cdot\Norm{\OptimalArm}_2}
    \right)\\
    &=
    \Phi\left(-
        \frac
        {\DivParam\Radius}
        {\Inflation\ExploreRatio\sqrt d}
        \cdot
        \sqrt{\frac
            {d\NormOp{\SymCovMatrix_t}}
            {\NormNuc{\SymCovMatrix_t}}
        }
    \right)\\
    &\geq
    \Phi(-1)\,,
\end{align*}
where we used the second and third equation in \cref{eq:well-posed-def}. This completes the proof.
\end{proof}

We are now ready to state our main result.
\begin{thm}
\label{thm:lin-ts-regret}
Let $\Worth_t(\cdot)$ be optimistic with parameter $\OptProb$ whenever $\ITypical_t=1$, $\ITypicalM_t=1$, and $\IWellPosed_t=1$. When \cref{as:optimal-diversity,as:optimal-random-friendly} hold, and $\Inner{\Arm,\Param}\in[-1,1]$, for all $\Arm\in\ArmSet_t$, almost surely. Then, we have that
\begin{align*}
    \sum_{t=1}^{T}\Inner[\big]{\Param,\OptimalArm_t}-\Inner[\big]{\Param,\ChosenArm_t}
    &\leq
    \frac{4(\RadiusParam+\RadiusWorth)\sqrt T}{\OptProb}
    \left(
    \sqrt{\Dim\log\left(1+\frac{\lambda T\ArmBound^2}{\Dim}\right)}
    +
    \sqrt{\log T}
    \right)
    +2\sum_{t=1}^{T}\Indicator(\Thinness(\SymCovMatrix_t)>\ThinnessBound)\,,
\end{align*}
with probability at least $1-\frac3T$.
\end{thm}
Before describing the proof, we state a direct corollary of \cref{thm:lin-ts-regret} and \cref{lem:lin-ts-optimism}.
\begin{cor}[\LinTS{} with smaller inflation]\label{cor:lin-ts-regret}
Consider  \LinTS{} with inflation parameter $\Inflation$ satisfying
\begin{align}
\Inflation=\frac{\DivParam\ThinnessBound}{\ExploreRatio}\cdot\frac{\Radius}{\sqrt{d}}~~~ \text{and} ~~~~\frac{\RadiusWorth}{\Inflation}=\LowerOrder{\sqrt{\log T}}\,.\label{eq:cor-lin-ts-inflation}
\end{align}
Then, when \cref{as:optimal-diversity,as:optimal-random-friendly}  hold, the regret is at most
\begin{align*}	
	\frac{4(\RadiusParam+\RadiusWorth)\sqrt T}{\Phi(-1)}
	\left(
	\sqrt{\Dim\log\left(1+\frac{\lambda T\ArmBound^2}{\Dim}\right)}
	+
	\sqrt{\log T}
	\right)
	+2\sum_{t=1}^{T}\Indicator(\Thinness(\SymCovMatrix_t)>\ThinnessBound)\,,
\end{align*}
with probability at least $1-\frac3T$.
\end{cor}
The implication of \cref{cor:lin-ts-regret} is that, under \cref{as:optimal-diversity,as:optimal-random-friendly}, one can circumvent the need for the $\sqrt{d}$ inflation factor in the posterior of the \LinTS{} algorithm, as the parameter $\Inflation$ grows only logarithmically with respect to $T$.

By utilizing both \cref{thm:lin-ts-regret} and \cref{lem:oful-optimism}, we can derive a similar corollary for OFUL.
\begin{cor}[OFUL with smaller confidence intervals]\label{cor:oful-regret}
	Consider a version of OFUL which is an instance of ROFUL with worth function $\Worth_t(\Arm)=\Inner{\Arm,\Estimator_t}+\RadiusWorth\Norm{\Arm}_{\SymCovMatrix_t}$ such that
	 $\RadiusWorth$ satisfies
	\[
	\RadiusWorth=\frac{\DivParam\ThinnessBound}{\ExploreRatio}\cdot\frac{\Radius}{\sqrt{d}}\,.
	\]
	Then, when \cref{as:optimal-diversity,as:optimal-random-friendly}  hold, the regret is at most
	\begin{align*}
4(\RadiusParam+\RadiusWorth)\sqrt T
		\left(
		\sqrt{\Dim\log\left(1+\frac{\lambda T\ArmBound^2}{\Dim}\right)}
		+
		\sqrt{\log T}
		\right)
		+2\sum_{t=1}^{T}\Indicator(\Thinness(\SymCovMatrix_t)>\ThinnessBound)\,,
	\end{align*}
	with probability at least $1-\frac3T$.
\end{cor}
\cref{cor:oful-regret} demonstrates that, under \cref{as:optimal-diversity,as:optimal-random-friendly}, it is possible to improve the performance of the OFUL algorithm by running it with smaller confidence sets that are reduced by a factor of $\sqrt{d}$, given that $\RadiusWorth$ only logarithmically depends on $T$. It is worth noting that while the reduced confidence sets do not result in better upper bounds due to $\RadiusParam$ remaining of order $\OrderLog{\sqrt{d}}$, they may lead to improved empirical performance of the OFUL algorithm.

\begin{proof}[Proof of  \cref{thm:lin-ts-regret}]
First, let $(\IOptim_t)_{t\in[T]}$ be the adapted sequence of Bernoulli random variables such that $\IOptim_t=1$ whenever $\ITypical_t=1$, $\ITypicalM_t=1$, $\IWellPosed_t=1$, and
\begin{align*}
    \sup_{\Arm\in\ArmSet_t}\Worth_t(\Arm)
    >
    \sup_{\Arm\in\ArmSet_t}\Inner{\Arm,\Param}
\end{align*}
simultaneously. It follows from the definition of optimism that $\Prob{\IOptim_t=1\Given\HistoryPlus_{t},\ArmSet_t}\geq\OptProb$. Let $t\in[T]$ be fixed and assume that $\ITypical_t=1$, $\ITypicalM_t=1$, $\IWellPosed_t=1$, and $\IOptim_t=1$. Then, we
\begin{align*}
    \Inner[\big]{\Param,\OptimalArm_t}-\Inner[\big]{\Param,\ChosenArm_t}
    &\leq
    \Worth_t(\ChosenArm_t)-\Inner[\big]{\Param,\ChosenArm_t}
    \\
    &\leq
    \ParenDelim*{\Inner[\big]{\Estimator_t,\ChosenArm_t}+\RadiusWorth\Norm{\ChosenArm_t}_{\SymCovMatrix_t}}
    -
    \ParenDelim*{\Inner[\big]{\Estimator_t,\ChosenArm_t}-\RadiusParam\Norm{\ChosenArm_t}_{\SymCovMatrix_t}}
    \\
    &\leq
    (\RadiusParam+\RadiusWorth)\Norm{\ChosenArm_t}_{\SymCovMatrix_t}\,.
\end{align*}
Furthermore, since $\Inner[\big]{\Param,\OptimalArm_t}-\Inner[\big]{\Param,\ChosenArm_t}\leq2$
and $\RadiusParam+\RadiusWorth\ge 2$, 
we have 
\[
 \Inner[\big]{\Param,\OptimalArm_t}-\Inner[\big]{\Param,\ChosenArm_t} \leq 2 \min\left(\frac{\RadiusParam+\RadiusWorth}{2}\Norm{\ChosenArm_t}_{\SymCovMatrix_t},1\right)\leq (\RadiusParam+\RadiusWorth)\min\left(\Norm{\ChosenArm_t}_{\SymCovMatrix_t},1\right)\,,
\]
and, 
\begin{align}
    \Expect*{
    Z_t
    \Given
    \HistoryPlus_t,\ArmSet_t
    }
    \leq
    0\,,
    \label{eq:super-mg-diff}
\end{align}
almost surely, where $Z_t$ is defined as

\begin{align*}
    Z_t
    :=
    \ITypical_t\ITypicalM_t\IWellPosed_t\left\{
    \Inner[\big]{\Param,\OptimalArm_t}
    -
    \Inner[\big]{\Param,\ChosenArm_t}
    -
    \frac{(\RadiusParam+\RadiusWorth)\IOptim_t}{\OptProb}
    \cdot
    \min\big(\Norm{\ChosenArm_t}_{\SymCovMatrix_t}, 1\big)
    \right\}.
\end{align*}
It follows from \cref{eq:super-mg-diff} that $\sum_{t=1}^{T}Z_t$ is a super-martingale, and noting that $\Abs{Z_t}\leq\frac{2(\RadiusParam+\RadiusWorth)}{\OptProb}$, it follows from Azuma's inequality that
\begin{align}
    \Prob*{
    \sum_{t=1}^{T}Z_t
    \geq
    \frac{4(\RadiusParam+\RadiusWorth)\sqrt{T\log T}}{\OptProb}
    }
    \leq
    \frac1T.
    \label{eq:super-mg-azuma}
\end{align}
Next, by applying the Cauchy-Schwartz inequality and Lemma 11 in \cite{abbasi2011improved}, we deduce that
\begin{align*}
    \sum_{t=1}^{T}\left(Z_t-\ITypical_t\ITypicalM_t\IWellPosed_t
        \Inner[\big]{\Param,\OptimalArm_t-\ChosenArm_t}\right)
    &\geq
    -\frac{(\RadiusParam+\RadiusWorth)}{\OptProb}
    \sum_{t=1}^{T}\min\big(\Norm{\ChosenArm_t}_{\SymCovMatrix_t}, 1\big)
    \\
    &\geq
    -\frac{(\RadiusParam+\RadiusWorth)}{\OptProb}
    \sqrt{T\sum_{t=1}^{T}\min\big(\Norm{\ChosenArm_t}_{\SymCovMatrix_t}^2, 1\big)}
    \\
    &\geq
    -\frac{(\RadiusParam+\RadiusWorth)}{\OptProb}
    \sqrt{2T\Dim\log\left(1+\frac{\lambda T\ArmBound^2}{\Dim}\right)}.
\end{align*}
By combining the above inequality with \cref{eq:super-mg-azuma}, we obtain
\begin{align}
    \Prob*{
    \sum_{t=1}^{T}\ITypical_t\ITypicalM_t\IWellPosed_t\Inner[\big]{\Param,\OptimalArm_t-\ChosenArm_t}
    \geq
    \frac{4(\RadiusParam+\RadiusWorth)\sqrt T}{\OptProb}
    \left(
    \sqrt{\Dim\log\left(1+\frac{\lambda T\ArmBound^2}{\Dim}\right)}
    +
    \sqrt{\log T}
    \right)
    }
    \leq
    \frac1T
    \label{eq:sum-tw-tail-bound}
\end{align}
We now turn to bounding $(1-\ITypical_t\ITypicalM_t\IWellPosed_t)\Inner[\big]{\Param,\OptimalArm_t-\ChosenArm_t}$. Notice that
\begin{align*}
    (1-\ITypical_t\ITypicalM_t\IWellPosed_t)\Inner[\big]{\Param,\OptimalArm_t-\ChosenArm_t}
    &\leq
    2(1-\ITypical_t\ITypicalM_t\IWellPosed_t)\\
    &=
    2(1-\ITypical_t\ITypicalM_t\IWellPosed_t)\Indicator(\Thinness(\SymCovMatrix_t)\leq\ThinnessBound)
    +
    2(1-\ITypical_t\ITypicalM_t\IWellPosed_t)\Indicator(\Thinness(\SymCovMatrix_t)>\ThinnessBound)\\
    &\leq
    2(1-\ITypical_t\ITypicalM_t\IWellPosed_t)\Indicator(\Thinness(\SymCovMatrix_t)\leq\ThinnessBound)
    +
    2\Indicator(\Thinness(\SymCovMatrix_t)>\ThinnessBound).
\end{align*}
Also, it follows from \cref{as:optimal-diversity,as:optimal-random-friendly} that for any $t\in[T]$
\begin{align*}
    \Prob*{
        \ITypical_t\ITypicalM_t\IWellPosed_t=0
        ~~~\text{and}~~~
        \Thinness(\SymCovMatrix_t)\leq\ThinnessBound
    }
    \leq
    \frac2{T^2}\,,
\end{align*}
which in combination with the union bound leads to
\begin{align*}
    \sum_{t=1}^{T}(1-\ITypical_t\ITypicalM_t\IWellPosed_t)\Inner[\big]{\Param,\OptimalArm_t-\ChosenArm_t}
    \leq
    2\sum_{t=1}^{T}\Indicator(\Thinness(\SymCovMatrix_t)>\ThinnessBound)
\end{align*}
with probability at least $1-\frac2{T}$. Finally, this together with \cref{eq:sum-tw-tail-bound} yield
\begin{align*}
    \sum_{t=1}^{T}\Inner[\big]{\Param,\OptimalArm_t-\ChosenArm_t}
    &\leq
    \frac{4(\RadiusParam+\RadiusWorth)\sqrt T}{\OptProb}
    \left(
    \sqrt{\Dim\log\left(1+\frac{\lambda T\ArmBound^2}{\Dim}\right)}
    +
    \sqrt{\log T}
    \right)+
    2
    \sum_{t=1}^{T}\Indicator(\Thinness(\SymCovMatrix_t)>\ThinnessBound)\,,
\end{align*}
with probability at least $1-\frac3T$.
\end{proof}
\section{Auxiliary Proofs}
\label{sec:auxi}
\begin{proof}[Proof of \cref{prop:bias-decomp}]
It is straightforward to see that $X_i | Y$ follows Gaussian distribution with mean $\frac{\sigma_i^2}{\sum_{i=1}^{n}\sigma_i^2}\cdot Y$. We thus get
\begin{align*}
    \Expect*{X_i\cdot g(Y,Z)}
    &=
    \Expect*{\Expect*{X_i\cdot g(Y,Z)\Given Y,Z}}\\
    &=
    \Expect*{\Expect*{X_i\Given Y,Z}\cdot g(Y,Z)}\\
    &=
    \frac{\sigma_i^2}{\sum_{i=1}^{n}\sigma_i^2}\cdot\Expect*{Y\cdot g(Y,Z)}\,.
\end{align*}
\end{proof}

\begin{proof}[Proof of \cref{lem:single-random-friendly}]
First of all, it follows from the definition of $\Arm$ that
\begin{align*}
    \Expect*{\Norm{\Arm}_{\SymCovMatrix_t}^2}=\frac1{3d}\Trace{\SymCovMatrix_t}=\frac1{3d}\NormNuc{\SymCovMatrix_t}.
\end{align*}
It then follows from the Hanson-Wright inequality (\textit{e.g.}, Theorem 6.2.1 in \cite{vershynin2018high}) that
\begin{align*}
    \Prob*{
        \Abs[\Big]{\Norm{\Arm}_{\SymCovMatrix}^2-\frac1{3\Dim}\NormNuc{\SymCovMatrix}}
        \leq
        \frac1{6\Dim}\NormNuc{\SymCovMatrix}
    }
    \leq
    \exp\left(-c\min\left\{
        \frac{\NormNuc{\SymCovMatrix_t}^2}{\NormF{\SymCovMatrix_t}^2},
        \frac{\NormNuc{\SymCovMatrix_t}}{\NormOp{\SymCovMatrix_t}}
    \right\}\right)
\end{align*}
for some constant $c>0$. Noting that
\begin{align*}
    \NormOp{\SymCovMatrix_t}\NormNuc{\SymCovMatrix_t}
    &\geq
    \NormF{\SymCovMatrix_t}^2,
\end{align*}
we can simplify the above tail bound to get
\begin{align*}
    \Prob*{
        \Abs[\Big]{\Norm{\Arm}_{\SymCovMatrix}^2-\frac1{3\Dim}\NormNuc{\SymCovMatrix}}
        \leq
        \frac1{6\Dim}\NormNuc{\SymCovMatrix}
    }
    \leq
    \exp\left(-c
        \frac{\NormNuc{\SymCovMatrix_t}}{\NormOp{\SymCovMatrix_t}}
    \right)
    \leq
    \exp\left(
        -\frac{c\Dim}{\ThinnessBound^2}
    \right).
\end{align*}
\end{proof}

\begin{proof}[Proof of \cref{lem:single-arm-diversity}]
Note that for all $s>0$ we have that
\begin{align*}
    \Expect{\exp\left(s\Inner{\Arm,V}\right)}
    &=
    \prod_{i=1}^{d}\Expect{\exp\left(s\Arm_iV_i\right)}\\
    &\leq
    \prod_{i=1}^{d}\exp\left(\frac{s^2V_i^2}{2\Dim}\right)\\
    &=
    \exp\left(\frac{s^2\sum_{i=1}^{d}V_i^2}{2\Dim}\right)\\
    &=
    \exp\left(\frac{s^2}{2\Dim}\right).
\end{align*}
It thus follows from the Chernoff bound that
\begin{align*}
    \Prob*{\Inner{\Arm,V} > \sqrt{\frac{2\log(1/p)}{\Dim}}}
    \leq
    p,
\end{align*}
which is the desired result.
\end{proof}

\end{APPENDICES}


\bibliography{papers,mypapers,books}

\end{document}